\pdfoutput=1
\documentclass[11pt]{article}
\usepackage[]{acl}
\usepackage{times}
\usepackage{colortbl} 
\usepackage{latexsym}
\usepackage[T1]{fontenc}
\usepackage[utf8]{inputenc}
\usepackage{microtype}
\usepackage{inconsolata}
\usepackage[addedmarkup=uwave,defaultcolor=red]{changes}
\usepackage{url}
\usepackage{tcolorbox}
\usepackage{graphicx}
\usepackage{makecell}
\usepackage{framed}
\usepackage{amsmath}
\usepackage{enumitem}
\usepackage{pifont}
\definecolor{shadecolor}{rgb}{0.92,0.92,0.92}
\usepackage{wrapfig}
\usepackage{booktabs}
\usepackage{amsfonts}       
\usepackage{nicefrac}       
\usepackage{microtype}      
\usepackage{xspace}
\usepackage{makecell}
\usepackage{multirow}
\usepackage[capitalize]{cleveref}
\usepackage{scalerel}
\usepackage[misc]{ifsym}
\usepackage{float}
\usepackage{stfloats}
\usepackage{ulem}
\usepackage{subfigure}
\usepackage{longtable}
\usepackage{makecell}
\makeatletter
\DeclareRobustCommand\onedot{\futurelet\@let@token\@onedot}
\def\@onedot{\ifx\@let@token.\else.\null\fi\xspace}

\def\ie{\emph{i.e}\onedot}

\makeatother

\makeatletter
\renewcommand{\paragraph}{%
  \@startsection{paragraph}{4}%
  {\z@}{0ex \@plus 0ex \@minus 0ex}{-1em}%
  {\normalfont\normalsize\bfseries}%
}
\makeatother

\newcommand{\method}{UAPO\xspace}
\newcommand{\DPO}{UAPO\xspace}
\newcommand{\SimPO}{SimUAPO\xspace}

\newcommand\blfootnote[1]{%
  \begingroup
  \renewcommand\thefootnote{}\footnote{#1}%
  \addtocounter{footnote}{-1}%
  \endgroup
}
\crefname{figure}{Figure}{Figures}
\crefname{table}{Table}{Tables}
\crefname{section}{Section}{Sections}

\title{Adaptive Preference Optimization with Uncertainty-aware Utility Anchor}

\author{
  Xiaobo Wang\textsuperscript{1,3},
  Zixia Jia\textsuperscript{3},
  Jiaqi Li\textsuperscript{3},
  Qi Liu\textsuperscript{1,2*},
  Zilong Zheng\textsuperscript{3*}
  \\
  \\
  \textsuperscript{1}State Key Laboratory of Cognitive Intelligence, University of Science and Technology of China\\
  \textsuperscript{2}Institute of Artificial Intelligence, Hefei Comprehensive National Science Center\\
  \textsuperscript{3}State Key Laboratory of General Artificial Intelligence, BIGAI
  \\
  \small{
   \texttt{xiaobow.0519@gmail.com, qiliuql@ustc.edu.cn,}
   }\\
   \small{
   \texttt{\{lijiaqi, jiazixia, zlzheng\}@bigai.ai}
   }
}

\begin{document}
\maketitle
\blfootnote{*Corresponding authors}

\begin{abstract}

Offline preference optimization methods are efficient for large language models (LLMs) alignment. Direct Preference optimization (DPO)-like learning, one of the most popular approaches, stands out for its efficiency in reward modeling. However, these methods typically follow the convention to use Bradley-Terry (BT) reward modeling that faces several critical assumptions, including the requirement for pairwise training data, model distribution shifting, human rationality assumption, etc. To address these limitations, we propose a general framework for offline preference optimization methods, Adaptive Preference Optimization with Utility Anchor (\method), which introduces an anchoring function to estimate the uncertainties brought from preference data annotation. Our method enables training even in scenarios where the data is unpaired, significantly enhancing data utilization efficiency. Moreover, the anchor design makes \method more robust in the training process. Experimental results demonstrate that \method achieves competitive outcomes without the strict dependency on data pairing, paving the way for more flexible and effective preference optimization methods.

\end{abstract}

\section{Introduction}

Aligning large language models (LLMs) with human values and preferences is a crucial step in LLM training. This alignment ensures that LLMs can better follow human instructions, becoming more helpful, harmless, and honest. However, ensuring that these models align with human preferences and ethical standards remains a significant hurdle. Previous work has made strides in this direction by employing techniques.
Reinforcement learning from human feedback (RLHF)~\cite{ouyang2022training} is a widely adopted method in the alignment domain, yielding significant improvements in LLM performance. However, the high computational cost and time consumption associated with RLHF present substantial challenges, limiting its broader application. To address these challenges, offline preference optimization methods, such as DPO~\cite{rafailov2023dpo}, have been developed as a more manageable offline training process. This shift makes the training procedure more controllable and resource-efficient.

Despite the inspiring progress and superior benchmark performance made, current methodologies predominantly adhere to the Bradley-Terry (BT; \citet{bradley1952rank}) model or its generalized form, the Plackett-Luce model~\cite{plackett1975analysis}, for reward modeling, a rank-based method widely applied to predict rational preference distribution given pairwise data. In this work, we challenge this convention in \cref{sec:limitations} by thoroughly analysis of its discrepancies with real-world scenarios. From the \textbf{data} perspective, the BT model's reliance on pairwise comparisons imposes a significant constraint on the collection of preference data, such as identifying winning and losing responses from a superior model or a pre-trained reward function. This process is both labor-intensive and data-inefficient. From the \textbf{optimization} perspective, over-optimization in preference learning, as highlighted by Goodhart's law~\cite{gao2023scaling}, can lead to distribution shifts and reward hacking when applied to out-of-distribution samples. This may also lead to a disparity between training and testing reward functions. From the \textbf{cognitive} perspective, the BT model presupposes that human annotators are fully rational and maximize utility values, which has been challenged by behavioral economics~\cite{Tversky1992} that humans are typically risk-averse. 
These limitations are also observed in synthetic data~\cite{cui2023ultrafeedback} generated by modern language models.

In this work, we introduce Adaptive Preference Optimization with Uncertainty-aware Utility Anchor (\textbf{\method}), a cognitively inspired preference optimization framework designed to address the above limitations across data, learning, and cognitive dimensions. 
Specifically, drawing inspiration from the anchoring effect identified in behavioral economics by \citet{simon1955behavioral}, \method introduces the concept of a \textit{utility anchor} to better model human preferences in a more flexible and robust manner to solve the uncertainty in the data labeling process. This integration allows for more efficient processing of unpaired data and yields a more precise estimation of human preferences.

By decoupling the learning objective into unpaired forms, the utility anchor enhances robustness to various hyperparameter settings, eliminating the necessity for meticulous customization or tuning across different models and datasets. 
 
We show that \method is flexible and can easily be integrated to enhance previous offline reward learning methods, such as DPO~\cite{rafailov2023dpo} and SimPO~\cite{meng2024simpo}.
Additionally, we provide theoretical analysis from the perspective of uncertainty penalty, a strategy commonly used in pessimistic RL that use uncertainty rewards~\cite{jin2021pessimism} to address the overoptimization problem in offline reward learning.

In \cref{sec:experiment}, we run extensive experiments to demonstate the effectiveness of our \method and its variant \SimPO. We highlight the main advantages of \method as follows.

\begin{itemize}[leftmargin=*, topsep=0pt, noitemsep]
    \item \textbf{A general framework for offline preference optimization:} \method can transform most offline preference optimization methods, which typically require pairwise data, into methods capable of effectively utilizing unpaired data. 
    \item \textbf{The utility anchor helps mitigate uncertainty in data labeling:} By naturally leveraging unpaired data, the utility anchor is inherently aware of uncertainty. In \method, it further guides the model to effectively handle ambiguous or inconsistent preference signals.
    \item \textbf{Rigorous theoretical analysis of utility anchor:} we prove that the utility anchor is more consistent with concurrent offline preference modeling, making the alignment process more robust.
\end{itemize}

Our results highlight the potential of \method as a versatile and robust solution for preference optimization, paving the way for its application in real-world scenarios where data and computational resources are limited.

\section{Uncertainty-aware Utility Anchor}

In this section, we start by briefly discussing prior offline preference learning methods ($\S$\ref{sec:background}). Then we thoroughly analyze critical assumptions of reward modeling in previous methods that do not hold in real-world settings  ($\S$\ref{sec:limitations}). To address the issues, we introduce a cognitive compelling utility anchor and devise our formulation of \method in $\S$\ref{sec:utility_anchor}. We further provide theoretical justification in $\S$\ref{sec:theory} from the perspective of pessimistic RL.

\subsection{Background: Offline Preference Learning}\label{sec:background}

Consider an instructed preference dataset $\mathcal{D} = \{x, y_w, y_l\}$, where $x$ represents the prompt, and $(y_w, y_l)$ are two responses generated by a reference model $\pi_{ref}$. $y_w$ signifies the preferred (winning) response, and $y_l$ denotes the dispreferred (losing) response, both of which can be annotated by human annotators or a language model. The function $r(x, y)$  is a latent reward function employed to model the preference comparison between $y_w$ and $y_l$ using the Bradley-Terry (BT) model ~\cite{bradley1952rank}, or its generalized form Plackett-Luce~\cite{plackett1975analysis}, given by:
\begin{equation}
\small
\begin{aligned}
    p(y_w \succ y_l \mid x) &= \frac{\exp\left(r(x, y_w)\right)}{\exp\left(r(x, y_w)\right) + \exp\left(r(x, y_l)\right)} \\
    &= \sigma (r(x, y_w) - r(x, y_l)),
    \label{eq:bt_model}
\end{aligned}
\end{equation}
where $\sigma(\cdot)$ denotes the sigmoid function.

\noindent\textbf{Direct Preference Optimization (DPO)}~\cite{rafailov2023dpo} is a leading offline preference optimization method that reparameterizes the reward function $r(x, y)$ using a closed-form expression with the optimal policy:
\begin{equation}
\small
r_{\text{DPO}}(x, y) = \beta \log  \frac{\pi_\theta(y | x)}{\pi_{\text{ref}}(y | x)}  + \beta \log Z(x),
\end{equation}
where $\pi_\theta$ is the policy model, $\pi_{\text{ref}}$ is the reference policy (typically the SFT model), and $Z(x)$ is the partition function. Integrating $r_{\text{DPO}}(x, y)$ into BT model yields the loss for DPO:
\begin{equation}
\small
\mathcal{L}_{\text{DPO}} = -\mathbb{E}_{(x, y_w, y_l) \sim \mathcal{D}} \left[ \log \sigma \left( r(x, y_w) - r(x, y_l) \right) \right].
\end{equation}

\noindent\textbf{Simple Preference Optimization (SimPO)}~\cite{meng2024simpo}, as a variant of DPO, demonstrates superior efficacy across diverse tasks by utilizing a reference-free reward with response length normalization:
\begin{equation}
\small
r_{\text{SimPO}}(x, y) = \frac{\beta}{|y|} \log  \pi_\theta(y | x),
\label{eq:simpo_reward}
\end{equation}
where $|y|$ is the length of response $y$. 
A target reward margin $\gamma$ is introduced to distinguish between the preference rewards ensuring the reward difference between winning and losing responses exceeds this margin. The loss of SimPO is as below:
\begin{equation}
\small
\begin{aligned}
\mathcal{L}_{\text{SimPO}} = &-\mathbb{E}_{(x, y_w, y_l) \sim \mathcal{D}} \Big[ 
\log \sigma \big( r_{\text{SimPO}}(x, y_w) \\ &- r_{\text{SimPO}}(x, y_l) - \gamma \big)  \Big].
\end{aligned}
\label{eq:simpo}
\end{equation}

\subsection{Limitations of offline preference optimization methods} \label{sec:limitations}

In this study, we delineate three critical issues from the perspectives of data, optimization, and cognition, that are typically treated as conventional in previous formulations of preference modeling (\cref{tab:align_methods}).

\begin{table}[t!]
\centering
\resizebox{\linewidth}{!}{
\begin{tabular}{@{}ll@{}}
\toprule
\textbf{Method} & \textbf{Objective} \\ 
\midrule
DPO  & $-\log \sigma \left( \beta \log \frac{\pi_\theta(y_w|x)}{\pi_{\text{ref}}(y_w|x)} - \beta \log \frac{\pi_\theta(y_l|x)}{\pi_{\text{ref}}(y_l|x)} \right)$ \\
\midrule
IPO  & $\left( \log \frac{\pi_\theta(y_w|x)}{\pi_{\text{ref}}(y_w|x)} - \log \frac{\pi_\theta(y_l|x)}{\pi_{\text{ref}}(y_l|x)} - \frac{1}{2\tau} \right)^2$ \\
\midrule
CPO  & $-\log \sigma \left( \beta \log \pi_\theta(y_w|x) - \beta \log \pi_\theta(y_l|x) \right) - \lambda \log \pi_\theta(y_w|x)$ \\
\midrule
\multirow{2}{*}{KTO}  & $-\lambda_w \sigma \left( \beta \log \frac{\pi_\theta(y_w|x)}{\pi_{\text{ref}}(y_w|x)} - z_{0} \right) + \lambda_l \sigma \left( z_{0} - \beta \log \frac{\pi_\theta(y_l|x)}{\pi_{\text{ref}}(y_l|x)} \right)$, \\ 
& where $z_{0} = \mathbb{E}_{(x, y) \sim \mathcal{D}} \left[\beta\text{KL}\left(\pi_\theta(y|x) \| \pi_{\text{ref}}(y|x)\right)\right]$ \\
\midrule
\multirow{2}{*}{ORPO}  & $-\log p_\theta(y_w|x) - \lambda \log \sigma \left( \frac{p_\theta(y_w|x)}{1 - p_\theta(y_w|x)} \right) - \log \left( \frac{p_\theta(y_l|x)}{1 - p_\theta(y_l|x)} \right)$, \\ 
& where $p_\theta(y|x) = \exp \left( \frac{1}{|y|} \log \pi_\theta(y|x) \right)$ \\
\midrule
R-DPO  & $-\log \sigma \left( \beta \log \frac{\pi_\theta(y_w|x)}{\pi_{\text{ref}}(y_w|x)} - \beta \log \frac{\pi_\theta(y_l|x)}{\pi_{\text{ref}}(y_l|x)} + \left(\alpha |y_w| - \alpha |y_l|\right) \right)$ \\
\midrule
SimPO & $-\log \sigma \left( \frac{\beta}{|y_w|} \log \pi_\theta(y_w|x) - \frac{\beta}{|y_l|} \log \pi_\theta(y_l|x) - \gamma \right)$ \\ 
\midrule
\midrule
\method & $-\log \sigma \Big( r(x, y_w) - r(x, y_{\bot}) \Big) - \log \sigma \Big( r(x, y_{\bot}) - r(x, y_l)\Big)$ \\ \bottomrule
\end{tabular}}
\caption{Comparison of different offline preference optimization methods and their objectives.}
\label{tab:align_methods}
\end{table}

\noindent\underline{\textit{Pairwise Training Data}}\quad{} From the \textbf{data} perspective, most preference optimization methods rely on pairwise comparison data to align a policy model \( \pi_\theta \), requiring multiple responses to be compared to determine alignment with human preferences. However, human preferences are often expressed without comparisons, such as \textit{stating a favorite fruit without indicating dislikes}. Such rationale also follows prior literature in prospect theory~\cite{Tversky1992}, later formalized as KTO~\cite{ethayarajh2024kto}, claiming that the reward function shall be a human-aware objective, in which they incorporate a pre-defined middle state $z_0$ to estimate the expected KL distance between $\pi_\theta(x_{data})$ and $\pi_{ref}(x_{data})$, which is nevertheless computationally intractable. Therefore, this estimation can only be approximated through sampling, which often leads to suboptimal performance.

\noindent\underline{\textit{Distribution Shifting}}\quad{} From the \textbf{optimization} perspective, it has been well studied the issue of over-optimization~\cite{gao2023scaling} in preference learning. 
In accordance with Goodhard's law, optimizing reward functions with specific in-distribution training data can inevitably lead to a distribution shift of outputs given an out-of-distribution~(OOD) sample, \ie, reward hacking. In this context, the OOD sample corresponds to real-world human preferences that deviate from those represented in the training distribution. Moreover, a discrepancy is created between training and testing reward functions: for training, a relative reward \textit{w.r.t.} losing sample $y_l$ is learned; and for testing, the policy $\pi_\theta$ has to generate a response solely based on input $x$ without relying on $y_l$ and reference model $\pi_{ref}$. This mismatch can cause the reward model to produce unreliable signals, thereby impairing the effectiveness of the policy during inference.

\noindent\underline{\textit{``Irrational'' Preference Pairs}}\quad{} From the \textbf{cognitive} perspective, a critical assumption of the Bradley-Terry (BT) model (\citet{bradley1952rank}; \cref{eq:bt_model}) is that annotators behave as fully rational agents who make pairwise comparisons by maximizing utility values with respect to an absolute, universal utility function~\cite{fisch2024robust,azar2024general}. However, this assumption often fails to hold in practice, as irrational behaviors are commonly observed, both in the preference data annotation process, where annotator judgments may be inconsistent or biased, and in the optimization stage, where model updates may not strictly follow utility-maximizing principles. Moreover, behavioral economists have identified that humans are \textit{risk averse}~\cite{Tversky1992}: decisions are made in \textit{relative} terms rather than absolutes. This phenomenon is also observed in modern language models~\cite{jia2024decision}, which are frequently employed as automated preference data generators~\cite{cui2023ultrafeedback} or evaluation judges~\cite{arenahard2024}. Such ``irrationality'' results in the modeling of moderate rewards rather than true utility functions, as presented in our analysis of reward comparisons (\cref{fig:threshold}).

\subsection{Reward Modeling with Utility Anchors}\label{sec:utility_anchor}

One critical challenge of the above issues is that the true distribution $p_{\text{human}} (y_w \succ y_l | x)$ is not known, while the observed data pairs $(x, y_w, y_l) \sim \mathcal{D}$ are subject to human aware uncertainties. In this section, we borrow the idea of ``reference anchors''~\cite{Tversky1992} and propose a more robust and flexible preference optimization framework in terms of data efficiency, reward shifting, and cognitive compelling.

\paragraph{Anchoring Effect} The anchoring effect, originally raised by behavior economist \citet{simon1955behavioral}, describes the common human tendency to rely heavily on a reference datapoint (the "anchor"), mostly the first information, when making decisions. Inspired by the cognitive bias, we curate a \textbf{utility anchor} that is better in line with concurrent offline preference modeling, defined as:
\begin{equation}
    \small
    r(x, y_{\bot}) = \beta\log \frac{\pi_\theta(x, y_\bot)}{\pi_{\text{ref}}(x, y_\bot)} + \beta \log Z(x) + \gamma,
    \label{eq:anchor_reward}
\end{equation}
where $y_\bot$ is a dummy token that can be learned through preference optimization, $\gamma$ is a constant reward anchor (more details can be seen in \cref{appendix:utility_anchor}). By introducing the utility anchor, the model gains the ability to handle uncertainty in preference data labeling (§\ref{sec:limitations}), allowing it to deal with ambiguous or inconsistent signals more effectively.

Since the sigmoid function is monotonically increasing, we have
\begin{equation}
\small
\begin{aligned}
        \log \sigma \left(r(x, y_w) - r(x, y_\bot) \right) + \log \sigma\left(r(x, y_\bot) - r(x, y_l)\right) \\ \leq \log \sigma\left(r(x, y_w) - r(x, y_l)\right).
\end{aligned}
\end{equation}

Thus, given the definition of utility anchor, we can rewrite the log-likelihood form of \cref{eq:bt_model} into
\begin{equation}
    \small
    \begin{aligned}
    \log p(y_w \succ y_l | x) = \log\sigma(r(x, y_w) - r(x, y_\bot)) \\+ \log\sigma ( r(x, y_\bot) - r(x, y_l)).
        \label{eq:atpo_reward}
    \end{aligned}
\end{equation}

\paragraph{The \method Objective} Finally, we obtain the \method objective:
\begin{equation}
    \small
    \begin{aligned}
&\mathcal{L}_{\text{\method}}(\pi_\theta) = -\mathbb{E}_{(x, y_w, y_l) \sim \mathcal{D}} \Big[ \log \sigma \Big( r(x, y_w) - r(x, y_{\bot}) \Big) \\
&\quad + \log \sigma \Big( r(x, y_{\bot}) - r(x, y_l) \Big) \Big],
\label{eq:single_loss}
    \end{aligned}
\end{equation}

where $r(x, y)$ denotes the reward function in any offline preference optimization method, and $r(x, y_\bot)$ represents the utility anchor corresponding to the adopted method. For example, when applying SimPO within this framework, the reference-free reward $r_{\text{\SimPO}}(x, y)$ is given in \cref{eq:simpo_reward}, and the utility anchor can be reformulated as:
\begin{equation}
    \small
    \begin{aligned}
        r_{\text{\SimPO}} (x, y_\bot) = \beta \log \pi_\theta (x, y_\bot) + \gamma.
        \label{eq:simuapo_reward}
    \end{aligned}
\end{equation}

\paragraph{Learning from unpaired data} Most preference datasets exhibit a strong win/lose imbalance, where dispreferred responses significantly outnumber preferred ones. For example, in UltraFeedback ~\cite{cui2023ultrafeedback}, each prompt is used to query multiple LLMs to generate four candidate responses, which are then annotated by GPT-4. Following previous works, the responses are categorized into one winning response and three losing responses, resulting in data imbalance.

To better solve the data imbalance problem and harness the capability of \method to effectively process unpaired data, we construct multiple datasets ($\S$\ref{sec:data}). And considering the separate form of \cref{eq:atpo_reward}, we could expand the objective function to accommodate multiple datasets. 
Given a set of $n$ responses $\{y_1, \cdots, y_n\}$ generated by different language models, we aim to induce an utility anchor hidden within human (or LLM) judges, \textit{s.t.} $(y_{w1} \succ y_{l1}), (y_{w2} \succ y_{l2}), \cdots$. Then the above objective turns into (refer to \cref{appendix:multi} for proof):
\begin{equation}
    \small
    \begin{aligned}
    &\mathcal{L}_w = -\sum_{y_w} \log \Big(\frac{\exp(r(x, y_w))}{\exp(r(x, y_\bot)) + \sum_{y_w'} \exp (r(x, y_w'))}\Big) \\
    &\mathcal{L}_l = -\log \Big(\frac{\exp(r(x, y_\bot))}{\exp(r(x, y_\bot)) + \sum_{y_l} \exp (r(x, y_l))}\Big) \\
    &\mathcal{L}_{{\textit{\method}\_multi} } =\mathbb{E}_{(x, y_w, y_l) \sim \mathcal{D}} \left(\mathcal{L}_{w} + \mathcal{L}_{l}\right).
    \end{aligned}
    \label{con:multi_loss}
\end{equation}

This adjustment allows \method to effectively exploit the availability of preferred and dispreferred responses without the need for explicitly paired counterparts.

Among the methods in \cref{tab:align_methods}, KTO also supports learning from unpaired data but is based on prospect theory, while \method is motivated by uncertainty penalties in reinforcement learning optimization. KTO relies on a fixed middle state $z_0$ to estimate the expected KL divergence between $\pi_\theta(x_{\text{data}})$ and $\pi_{\text{ref}}(x_{\text{data}})$, which is approximated via prompt sampling and often underperforms. In contrast, \method learns a utility anchor through a trainable linear representation, offering better efficiency and performance.

\subsection{Theoretical Analysis: Preference Optimization with Uncertainty Penalty}\label{sec:theory}

From the optimization objective, we provide further interpretation of the \method from the view of pessimistic RL, and justify how it addresses the distribution shifting given the OOD samples ($\S$\ref{sec:limitations}).

The utility anchor defined in \cref{eq:anchor_reward} can be rewritten into an output-independent score:
\begin{equation}
    \small
    r_\phi(x) = \beta 
\log \frac{u_\phi(\pi_\theta(x))}{u_\phi(\pi_
    {\text{ref}}(x))} + \beta \log Z(x) +\gamma,
\end{equation}
where $u$ is a linear function parameterized by $\phi$ to model the utility anchor $y_\bot$. Applying this to the \method objective in \cref{eq:single_loss} and set $\gamma=0$ for simplicity and fair theoretical comparison with DPO:
\begin{equation}
    \small
    \begin{aligned}
\max_{\pi_\theta, u_\phi} \mathbb{E} [\sigma(\beta\log \frac{\pi_\theta (x, y_w)}{\pi_{\text{ref}}(x, y_w)} - \beta \log \frac{u_\phi(\pi_\theta(x))}{u_\phi(\pi_
    {\text{ref}}(x))}) \\
    +\sigma( \beta \log \frac{u_\phi(\pi_\theta(x))}{u_\phi(\pi_
    {\text{ref}}(x))} - \beta\log \frac{\pi_\theta (x, y_l)}{\pi_{\text{ref}}(x, y_l)})
    ].
    \end{aligned}
\end{equation}

Given a learned $u_\phi$, the derivative of the preferred objective $\mathcal{L}_w$ (we skip the symmetric form of $\mathcal{L}_l$ for simplicity) is
\begin{equation}
    \small
    \begin{aligned}
    &\nabla \mathcal{L}_w =
    -\beta\mathbb{E} \Big[ \sigma(-(\underbrace{\hat{r}_\theta(x, y_w) - \textcolor{red}{\hat{u}_{\phi,\theta}(x)}}_{\text{uncertainty normalized reward}})) \\&[\nabla_\theta \log \pi(y_w|x) - \nabla_\theta \log u(\pi(x))] \Big],
    \end{aligned}
\end{equation}
where $\hat{r}_\theta(x, y) = \beta\log\pi(x,y) / \pi_{\text{ref}}(x, y)$ and $\hat{u}(x) = \beta \log u_\phi(\pi_\theta (x)) /  u_\phi (\pi _{\text{ref}}(x))$ can be considered a \textcolor{red}{uncertainty reward penalty}. Considering the increasing feature of the Sigmoid function, the objective is equivalent to optimizing the lower-bound of an uncertainty-aware policy likelihood (refer to \cref{appendix:elbo} for proof):
\begin{equation}
    \small
    \log \pi_{\theta,\phi}(y|x) = \log \frac{1}{Z(x)} \pi_{\text{ref}} (y|x) e^{\frac{1}{\beta} (r(x, y) - \textcolor{red}{\hat{u}(x)})},
    \label{eq:confidence_bound}
\end{equation}
where $\hat{u}(x)$ indicates the reference anchor of uncertainties given input $x$. The representation is also consistent with conventional pessimistic RL theories~\cite{jin2021pessimism} that produce a conservative estimate of reward as a lower confidence bound.

\section{Experiments}\label{sec:experiment}

\subsection{Setup}

\paragraph{Models and Training Dataset} \label{sec:data}

We select the instruct models of Mistral-7B ~\cite{jiang2023mistral}, Llama-3-8B ~\cite{llama3modelcard}, and the base model of Gemma-2-9B ~\cite{gemmateam2024gemma2improvingopen} for evaluation to acquire better instruction-following capabilities in this paper. 

We use three datasets including princeton-nlp/mistral-instruct-ultrafeedback \footnote{https://huggingface.co/datasets/princeton-nlp/mistral-instruct-ultrafeedback}, princeton-nlp/llama3-ultrafeedback-armorm \footnote{https://huggingface.co/datasets/princeton-nlp/llama3-ultrafeedback-armorm}, and princeton-nlp/gemma2-ultrafeedback-armorm \footnote{https://huggingface.co/datasets/princeton-nlp/gemma2-ultrafeedback-armorm} for training Mistral-7B, Llama-3-8B and Gemma-2-9B respectively. It is worth mentioned that we further reconstruct the training data in these three datasets to obtain one winning response companied with three losing responses, rather than one pair of winning and losing response for comparison. More details about \textbf{multiple datasets} construction can be seen in \cref{appendix:multi_data}.

\paragraph{Baselines}

We compare \method with various offline preference optimization methods including DPO ~\cite{rafailov2023dpo}, IPO ~\cite{azar2024general}, CPO ~\cite{xu2024contrastive}, KTO ~\cite{ethayarajh2024kto}, ORPO ~\cite{hong2024orpo}, R-DPO ~\cite{DBLP:conf/acl/rdpo} and SimPO ~\cite{meng2024simpo}. Each method addresses distinct challenges in aligning policy models with human preferences.

\paragraph{Benchmarks and Evaluation Metrics}

We evaluate our models on three widely recognized open-ended instruction-following benchmarks: AlpacaEval 2~\cite{alpaca_eval}, Arena-Hard v0.1~\cite{li2024crowdsourced, arenahard2024}, and MT-Bench~\cite{DBLP:conf/nips/mtbench}. These benchmarks are considered representative of human preference evaluations relative to the training data. AlpacaEval 2 consists of 805 questions sampled from five datasets, covering various instruction-following tasks. MT-Bench includes 80 questions distributed across eight categories, providing a structured evaluation of model performance on different conversational and reasoning tasks. Arena-Hard v0.1  extends MT-Bench by introducing 500 well-defined technical problem-solving questions, further challenging the models' reasoning and problem-solving abilities.

We consistently follow the usage of the evaluation metrics that are used in the original benchmark for each benchmark. For AlpacaEval 2, we present both the raw win rate (\textbf{WR}) and the length-controlled win rate (\textbf{LC})~\cite{dubois2024length} which is designed to mitigate the influence of model verbosity on preference judgments. For Arena-Hard, we use WR relative to a baseline model, allowing direct comparison of alignment performance. For MT-Bench, we compute the average score using GPT-4 as the judge model for assessing the instruction-following quality. 

We also evaluate our models on RewardBench ~\cite{rewardbench} and RewardBench 2 ~\cite{rewardbench2} to compare performance in the OOD settings. RewardBench and RewardBench 2 are widely known as comprehensive benchmarks for evaluating reward models, specifically focusing on preference alignment and generalization ability beyond training distributions. RewardBench provides a diverse set of prompts and human preference annotations designed to capture realistic deployment scenarios, while RewardBench 2 extends this setup with a broader and more challenging collection of tasks that place greater emphasis on consistency and robustness across domains. Together, these two benchmarks capture real-world human preferences and enable us to assess how well our models adapt to distribution shifts between training data and practical human evaluations.

\subsection{Main Results}

\paragraph{\method gains an improvement over existing preference optimization methods on pairwise datasets}

\begin{table*}[t!]
    \centering
    \resizebox{\linewidth}{!}{
    \begin{tabular}{lcccccccccccc}
        \toprule
        \multirow{3}{*}{\textbf{Method}} & \multicolumn{4}{c}{\textbf{Mistral-7B-Instruct}} & \multicolumn{4}{c}{\textbf{Llama-3-8B-Instruct}} & \multicolumn{4}{c}{\textbf{Gemma-2-9B-Instruct}} \\
        \cmidrule(lr){2-5} \cmidrule(lr){6-9} \cmidrule(lr){10-13}
        & \multicolumn{2}{c}{\textbf{AlpacaEval 2}} & \textbf{Arena-Hard} & \textbf{MT-Bench} & \multicolumn{2}{c}{\textbf{AlpacaEval 2}} & \textbf{Arena-Hard} & \textbf{MT-Bench} & \multicolumn{2}{c}{\textbf{AlpacaEval 2}} & \textbf{Arena-Hard} & \textbf{MT-Bench} \\
        \cmidrule(lr){2-3} \cmidrule(lr){4-4} \cmidrule(lr){5-5} \cmidrule(lr){6-7} \cmidrule(lr){8-8} \cmidrule(lr){9-9} \cmidrule(lr){10-11} \cmidrule(lr){12-12} \cmidrule(lr){13-13}
        & LC (\%) & WR (\%) & WR (\%) & GPT-4 &  LC (\%) & WR (\%) & WR (\%) & GPT-4 &  LC (\%) & WR (\%) & WR (\%) & GPT-4\\
        \midrule
        \multicolumn{13}{c}{\cellcolor{lightgray!50} \textbf{pairwise datasets}}\\
        SFT   & 19.0  & 15.4  & 12.9 & 7.5 & 26.0  & 25.3  & 22.3 & 8.1 & 48.7  & 36.5  & 42.1 & 8.6\\
        IPO   & 20.3  & 20.3  & 16.2 & \textbf{7.8} & 46.8  & 42.4  & \textbf{36.6} & \uline{8.2} & 62.6  & 58.4  & 53.5 & 8.7\\
        CPO   & 23.8  & 28.8  & \textbf{22.6} & 7.5 & 34.1  & 36.4  & 30.9 & \uline{8.2} & 56.4  & 53.4  & 55.2 & 8.7\\
        KTO   & 24.5  & 23.6  & 17.9 & \uline{7.7} & 34.1  & 32.1  & 27.3 & \uline{8.2} & 61.7  & 55.5  & 53.8 & 8.5\\
        ORPO  & 24.5  & 24.9  & \uline{20.8} & \uline{7.7} & 38.1  & 33.8  & 28.2 & \textbf{8.3} & 56.2  & 46.7  & 46.2 & 8.3\\
        R-DPO & 27.3  & 24.5  & 16.1 & 7.5 & 48.0  & \uline{45.8}  & \uline{35.1} & 8.2 & 68.3  & 66.9  & 57.9 & 8.5\\
        \midrule
         \textcolor{gray}{DPO$^\star$}   &  \textcolor{gray}{26.8} &  \textcolor{gray}{24.9}  &  \textcolor{gray}{16.3} & \textcolor{gray}{7.6} &  \textcolor{gray}{48.2} &  \textcolor{gray}{47.5} & \textcolor{gray}{35.2} & \textcolor{gray}{8.2} & \textcolor{gray}{67.8} & \textcolor{gray}{65.4} & \textcolor{gray}{58.9} & \textcolor{gray}{-}\\
        DPO   &  20.5  & 18.1  & 13.4 & 7.6 &  41.2 &  37.5 & 33.9 & \uline{8.2} & 68.8 & 64.5 & 58.4 & 8.7 \\
        \textcolor{gray}{SimPO$^\star$} & \textcolor{gray}{32.1}  &  \textcolor{gray}{34.8} & \textcolor{gray}{21.0} & \textcolor{gray}{7.6} & \textcolor{gray}{53.7}  & \textcolor{gray}{47.5}  & \textcolor{gray}{36.5} & \textcolor{gray}{8.0} & \textcolor{gray}{72.4}  & \textcolor{gray}{65.9}  & \textcolor{gray}{57.8} & \textcolor{gray}{-} \\
        SimPO &  \textbf{29.0} & \uline{31.9}  & 19.9 & 7.2 & \uline{50.1}  & 45.2  & 27.7 & 7.4 & \uline{72.6} & 66.1 & 57.8 & \uline{8.8} \\
        \midrule
        \DPO & 23.1  & 17.7 & 11.1 & \uline{7.7} & 41.2 & 38.3 & 32.1 & \uline{8.2} & 70.2  & \textbf{67.7}  & \uline{58.8} & \textbf{8.9}\\
        \SimPO & \uline{28.6}  & \textbf{32.2} & 17.7 & 7.6  & \textbf{51.2} & \textbf{47.6} &  34.4 & \textbf{8.3} & \textbf{73.5}  & \uline{67.0} & \textbf{59.4} & \textbf{8.9} \\
        \bottomrule
    \end{tabular}}
    \caption{Performance comparison of different methods on Mistral-Instruct, Llama-3-Instruct, and Gemma-2-Instruct trained with pairwise datasets. $\star$ denotes results reported by \citet{meng2024simpo}. The performance differences of DPO and SimPO are attributed to the version of GPT used.  The best results are highlighted in \textbf{bold}, while the second-best results are \uline{underlined}.}
    \label{tab:main_result}
\end{table*}

As shown in \cref{tab:main_result}, all offline preference optimization methods outperform the SFT baseline, demonstrating the effectiveness of preference-based training. Among them, \DPO and \SimPO consistently achieve superior results on Llama-3-Instruct and Gemma-2-Instruct compared to their respective counterparts, DPO and SimPO. Specifically, on AlpacaEval 2, \DPO outperforms DPO by 2.6 points on Mistral-Instruct and 1.4 on Gemma-2-Instruct, showing stronger alignment with human preferences. \SimPO achieves state-of-the-art LC results on Gemma-2-Instruct. On Arena-Hard, it improves over SimPO by 6.7 points on Llama-3-Instruct and 1.6 on Gemma-2-Instruct.

Meanwhile, we observe that MT-Bench scores are relatively consistent across the three models and all preference optimization methods. This may be due to the inherent randomness in MT-Bench scoring, as discussed in \cref{appendix:mt-bench}. While \SimPO performs comparably to SimPO on Mistral-Instruct and shows a slight drop in LC on AlpacaEval 2, this could be attributed to training dynamics specific to Mistral-Instruct.

In general, the consistently strong results across different models and benchmarks show the effectiveness of \method in preference-based alignment.

\paragraph{Multiple datasets can improve \method much more}

\begin{table}[t!]
    \centering
    \scalebox{0.65}{
    \begin{tabular}{lcccc}
        \toprule
        \multirow{2}{*}{\textbf{Method}} & \multicolumn{2}{c}{\textbf{AlpacaEval 2}} & \textbf{Arena-Hard} & \textbf{MT-Bench} \\
        \cmidrule(lr){2-3} \cmidrule(lr){4-4} \cmidrule(lr){5-5} 
        & LC (\%) & WR (\%) & WR (\%) & GPT-4 \\
        \midrule
        \multicolumn{5}{c}{\cellcolor{lightgray!50} \textbf{multiple datasets}}\\
        DPO   & 41.2  & 37.5  & 33.9 &  8.2 \\
        SimPO & 50.1  & 45.2  & 27.7 &  7.4 \\ 
        \DPO & 41.2 & 38.3 & 32.1 & 8.2 \\
        \SimPO & \uline{51.2} & \textbf{47.6} & \uline{34.4} & \uline{8.3} \\
        \midrule
        DPO-multi & 43.3 & 38.0 & 30.1 & 7.2 \\
        SimPO-multi & 45.7 & 41.6 & 27.5 & 6.7 \\
        \DPO-multi & 44.0 & 43.0 & 33.5 & 7.3 \\
        \SimPO-multi & \textbf{55.2} & \uline{47.2} & \textbf{35.2} & \textbf{8.4} \\
        \bottomrule
    \end{tabular}}
    \caption{Performance comparison of different methods on Llama-3-Instruct trained with multiple datasets.}
    \label{tab:llama-multi}
\end{table}

We utilize the multiple datasets mentioned in \cref{sec:data} to train DPO, SimPO, \DPO and \SimPO on Llama-3-Instruct. In \cref{tab:llama-multi}, DPO-multi, SimPO-multi, \DPO-multi, and \SimPO-multi refer to the respective training results using multiple datasets. And we set constant reward anchor $\gamma = 0$ in \DPO-multi training process. Compared to using pairwise datasets, \DPO and \SimPO show a significant improvement in training on multiple datasets.

Overall, \DPO and \SimPO demonstrate superior performance compared to DPO and SimPO when utilizing multiple datasets. Specifically, on Llama-3-Instruct, \SimPO-multi achieves a 55.2 LC on AlpacaEval 2, surpassing other offline preference optimization methods. However, the MT-Bench scores remain inconsistent, as their trends do not align with those observed on other benchmarks, further supporting our earlier observation regarding the inherent randomness in MT-Bench evaluations. 

Meanwhile, the performance of SimPO-multi lags behind that of SimPO, indicating that SimPO may not benefit from the multiple datasets. One possible reason is the lack of a reference model during training, which increases the model’s susceptibility to the inherent dispersion in the distribution of multiple datasets, ultimately leading to a less stable learned policy. Overall, these improvements suggest that by leveraging multiple datasets, \DPO-multi and \SimPO-multi are able to learn a more accurate reference anchor that can extract deeper information of the prompt, for classifying responses as either winning or losing. The enhanced anchor allows the policy model to better capture and understand the preferences of the data. 

Notably, \method achieves the best overall performance while also reducing training time. Unlike DPO and SimPO, which require three separate backward passes for three data pairs, \method uses a single pass, making it significantly more efficient.

\begin{figure*}[t!]
  \centering 
  \subfigure[Training Rewards and Utility Anchor]{ 
    \includegraphics[width=0.31\linewidth]{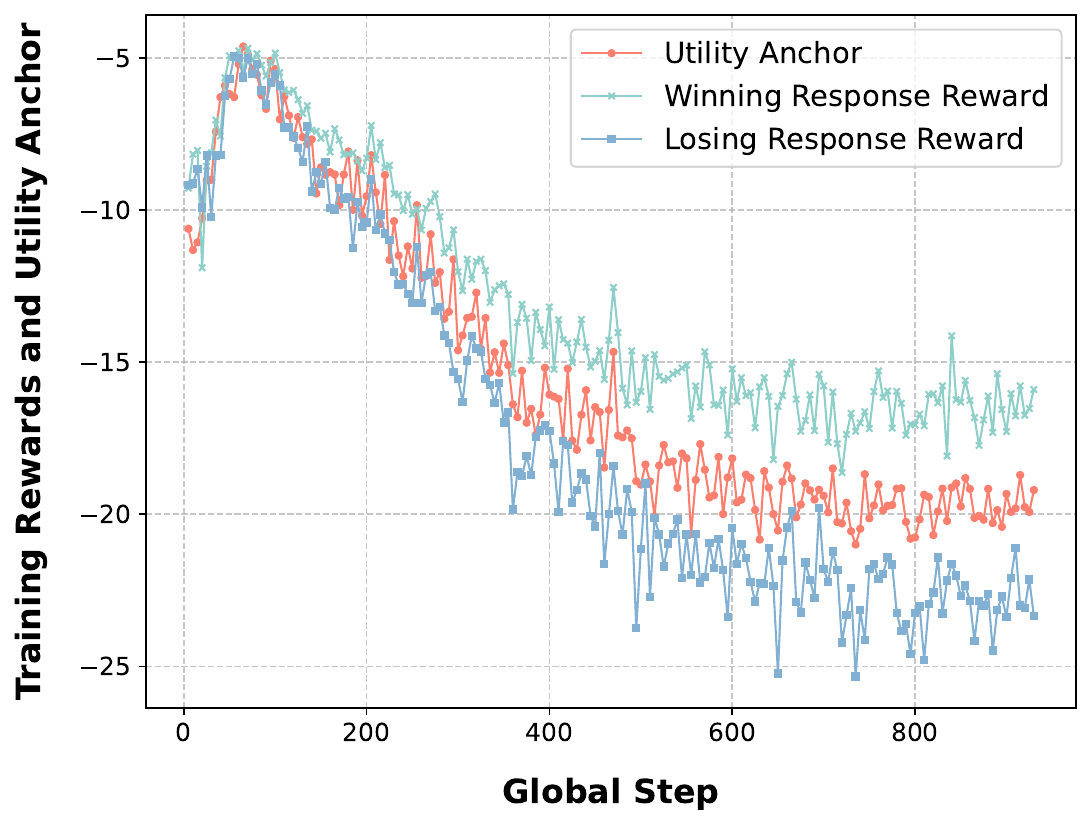} 
    \label{fig:threshold}
  } 
  \subfigure[Training Reward Margin]{ 
    \includegraphics[width=0.31\linewidth]{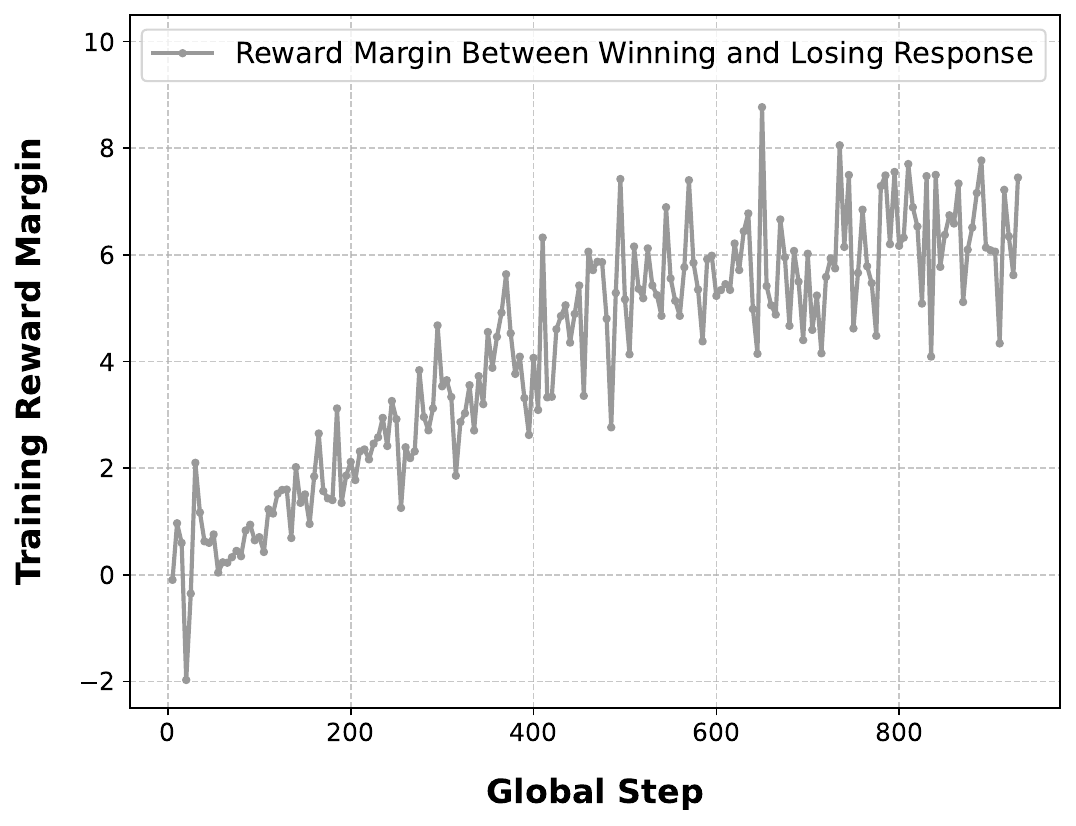}
    \label{fig:margin}
  } 
  \subfigure[Training Accuary]{ 
    \includegraphics[width=0.31\linewidth]{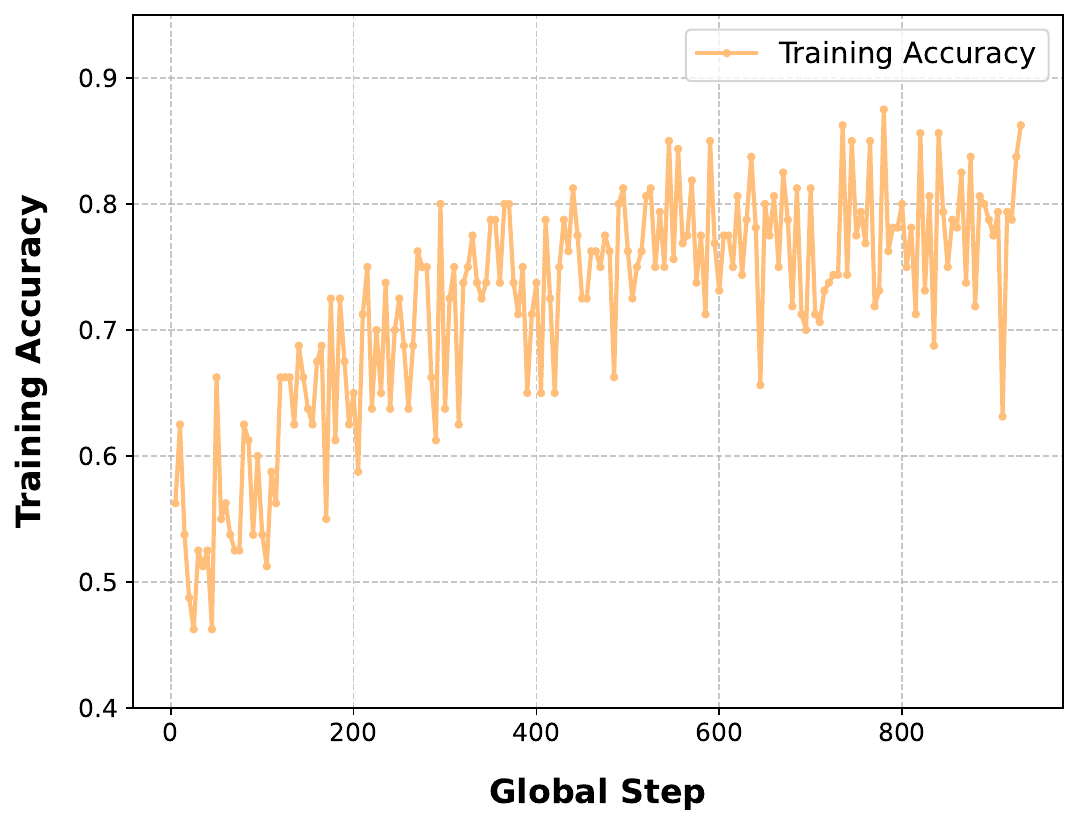}
    \label{fig:acc}
  } 
  \caption{Analysis of utility anchor representation during \SimPO training on Gemma-2-Instruct: (a) Relationship between the margin of winning and losing response reward; (b) Reward margin during training, and (c) Accuracy for training, i.e., the percentage of response rewards for winning greater than those of losing.} 
  \label{fig:training_curve}
\end{figure*}

\begin{table*}[t!]
    \centering
    \scalebox{0.6}{
    \begin{tabular}{lccc}
        \toprule
       \textbf{Prompt} & \textbf{Reward of $y_w$} & \textbf{Utility Anchor} & \textbf{Reward of $y_l$} \\
        \midrule
        Prompt A: \textit{Teacher:In this task you will be given a list of integers. You should remove any integer that is not prime ...} & -9.875 & -10.764 & -11.587 \\
        Prompt B: \textit{explain all machine learning algorithms with examples and code in python}  & -11.346 & -19.879 & -21.724 \\
        \bottomrule
    \end{tabular}}
    \caption{The intuitive example cases of Gemma-2-Instruct trained by \SimPO demonstrate that the utility anchor is closely related to the prompt. The reward of \( y_w \) for Prompt B is lower than the utility anchor of Prompt A. Therefore, if a constant standard is used, \( y_w \) from Prompt B will be classified as part of the losing responses.}
    \label{tab:prompt}
    \vspace{-8pt}
\end{table*}

\subsection{In-depth Analysis of \method}

\paragraph{The utility anchor effectively classifies responses of either winning or losing that closely relate to the prompt}

As one of the core contributions of this work, we introduce the utility anchor to offline preference optimization. To demonstrate the classification performance of the utility anchor, we present the training curve of \SimPO on Gemma-2-Instruct in \cref{fig:training_curve}. \cref{fig:threshold} illustrates a visualization that the utility anchor lies between the reward of the winning response and the reward of the losing response, indicating that the anchor effectively classifies responses as either winning or losing. Furthermore, the utility anchor ensures that the reward for the winning response decreases at a slower rate compared to the reward for the losing response. As a result, the reward margin, shown in \cref{fig:margin}, increases during the training process. Additionally, the training accuracy, depicted in \cref{fig:acc}, increases rapidly, as one of the anchor's effects is to enhance the accuracy of training

Since the utility anchor is based on the prompt, we also sample two intuitive examples to measure the degree of relativity between the prompt and the utility anchor. As shown in \cref{tab:prompt}, we select two prompts and present the rewards corresponding to the winning and losing responses for both the prompt and the utility anchor. The results reveal that the reward for the winning response of Prompt B is lower than the reference anchor for Prompt A. This underscores the importance of setting a utility anchor based on the prompt, as a static anchor would lead to incorrect classification.

\paragraph{Superior Generalization of \method under OOD Settings on RewardBench and RewardBench 2} As shown in \cref{tab:rewardbench}, \DPO and \SimPO consistently outperform their respective baselines, DPO and SimPO, across multiple evaluation dimensions. On RewardBench, our methods exhibit stronger overall alignment quality, with notable improvements in more challenging aspects such as reasoning and safety. On RewardBench 2, which introduces more challenging tasks such as factuality, mathematical reasoning, and precise instruction following, \DPO and \SimPO maintain a clear performance advantage. This indicates enhanced adaptability to distribution shifts and greater robustness in safety-critical evaluations.

These results demonstrate that \DPO not only improves preference alignment on standard benchmarks but also generalizes effectively to unseen scenarios, underscoring their robustness and practical applicability.

\begin{table*}[t!]
    \centering
    \resizebox{\linewidth}{!}{
    \begin{tabular}{lcccccccccccc}
        \toprule
        \multirow{2}{*}{\textbf{Method}} & \multicolumn{5}{c}{\textbf{RewardBench}} & \multicolumn{7}{c}{\textbf{RewardBench 2}} \\
        \cmidrule(lr){2-6} \cmidrule(lr){7-13}
        & Score & Chat & Chat Hard & Safety & Reasoning & Score & Factuality & Precise IF & Math & Safety & Focus & Ties \\
        \midrule
        \multicolumn{13}{c}{\cellcolor{lightgray!50} \textbf{Gemma-2-9B-Instruct}}\\
        DPO  & \uline{81.0} & 93.6 & 63.2 & 83.1 & 84.0 & 43.1 & 40.4 & 30.6 & 42.8 & 55.1 & 69.2 & 20.7  \\
        SimPO & 79.2 & 93.3  & 62.3 & 80.1 & 81.2 & 41.7 & 41.5 & 25.8 & 42.6 & 50.2 & 65.1 & 24.8 \\ 
        \DPO & \textbf{81.3} & 93.8  & 65.4 & 84.4 & 81.5 & \textbf{45.4} & 41.7 & 29.4 & 43.0 & 53.2 & 68.9 & 36.1 \\
        \SimPO & 80.8 & 93.6 & 61.3 & 86.1 & 82.0 & \uline{44.0} & 42.2 & 25.6 & 43.3 & 62.1 & 66.3 & 24.2 \\
        \midrule
        \multicolumn{13}{c}{\cellcolor{lightgray!50} \textbf{Llama-3-8B-Instruct}}\\
        DPO & 70.0 & 93.3 & 41.7 & 75.5 & 68.9 & 32.2 & 26.9 & 30.9 & 39.3 & 48.2 & 28.2 & 19.6 \\
        SimPO & \uline{72.5} & 93.0 & 53.1 & 74.2 & 69.8 & \uline{35.8} & 32.3 & 30.0 & 43.2 & 50.9 & 36.3 & 22.1 \\
        \DPO & 70.7 & 94.4 & 42.6 & 75.7 & 70.0 & 33.0 & 26.9 & 33.0 & 39.4 & 49.9 & 29.3 & 19.4 \\
        \SimPO & \textbf{72.7} & 93.9 & 53.2  & 75.3 & 68.5 & \textbf{36.2} & 33.2 & 30.1 & 38.9 & 51.0 & 33.8 & 29.9 \\
        \bottomrule
    \end{tabular}}
    \caption{Results on out-of-distribution reward modeling benchmarks.}
    \label{tab:rewardbench}
\end{table*}

\paragraph{\method exhibits superior robustness under preference distribution shifts between training and evaluation} To assess the robustness of preference optimization under distributional shifts, which commonly occur when aligning models with real-world human preferences, we simulate an out-of-distribution (OOD) training setting by randomly swapping the winning and losing responses in 40\% of the training samples. This synthetic noise introduces preference corruption that reflects potential misalignment between training distributions and true human values ($\S$\ref{sec:limitations}). We fine-tune both Mistral-Instruct and Llama-3-Instruct on this OOD dataset, and present the results in \cref{tab:OOD}. The performance of baseline methods such as SimPO and \SimPO declines due to the presence of corrupted preference signals. In contrast, \SimPO shows a smaller performance drop, suggesting that it is more robust to noisy preferences. We attribute this improvement to the utility anchor mechanism, which encourages the model to focus on the inherent semantics of the prompt rather than relying heavily on uncertain or inconsistent human feedback.

\begin{table}[h]
    \centering
    \scalebox{0.63}{
    \begin{tabular}{lcccc}
        \toprule
        \multirow{2}{*}{\textbf{Method}} & \multicolumn{2}{c}{\textbf{Mistral-7B-Instruct}} & \multicolumn{2}{c}{\textbf{Llama-3-8B-Instruct}} \\
        \cmidrule(lr){2-3} \cmidrule(lr){4-5}
        \cmidrule(lr){2-3} \cmidrule(lr){4-5}
        & LC (\%) & WR (\%) & LC (\%) & WR (\%) \\
        \midrule
        SimPO & 26.4 (-9.0\%) & 27.2 (-14.7\%) & 46.9 (-6.4\%) & \textbf{43.9 (-2.9\%)} \\
        \SimPO & \textbf{28.0 (-2.1\%)} & \textbf{30.9 (-4.0\%)} & \textbf{48.7 (-4.9\%)} & 41.8 (-12.2\%) \\
        \bottomrule
    \end{tabular}}
    \caption{Performance comparison of Mistral-Instruct and Llama-3-Instruct on AlpacaEval 2 under a distribution shift in preferences between training and evaluation data. Values in parentheses represent the relative performance degradation compared to models trained on the original training data.}
    \label{tab:OOD}
\end{table}

\subsection{KL divergence control in \method}

\begin{figure}[t!]
  \centering 
  \subfigure[DPO vs. \DPO]{ 
    \includegraphics[width=0.45\linewidth]{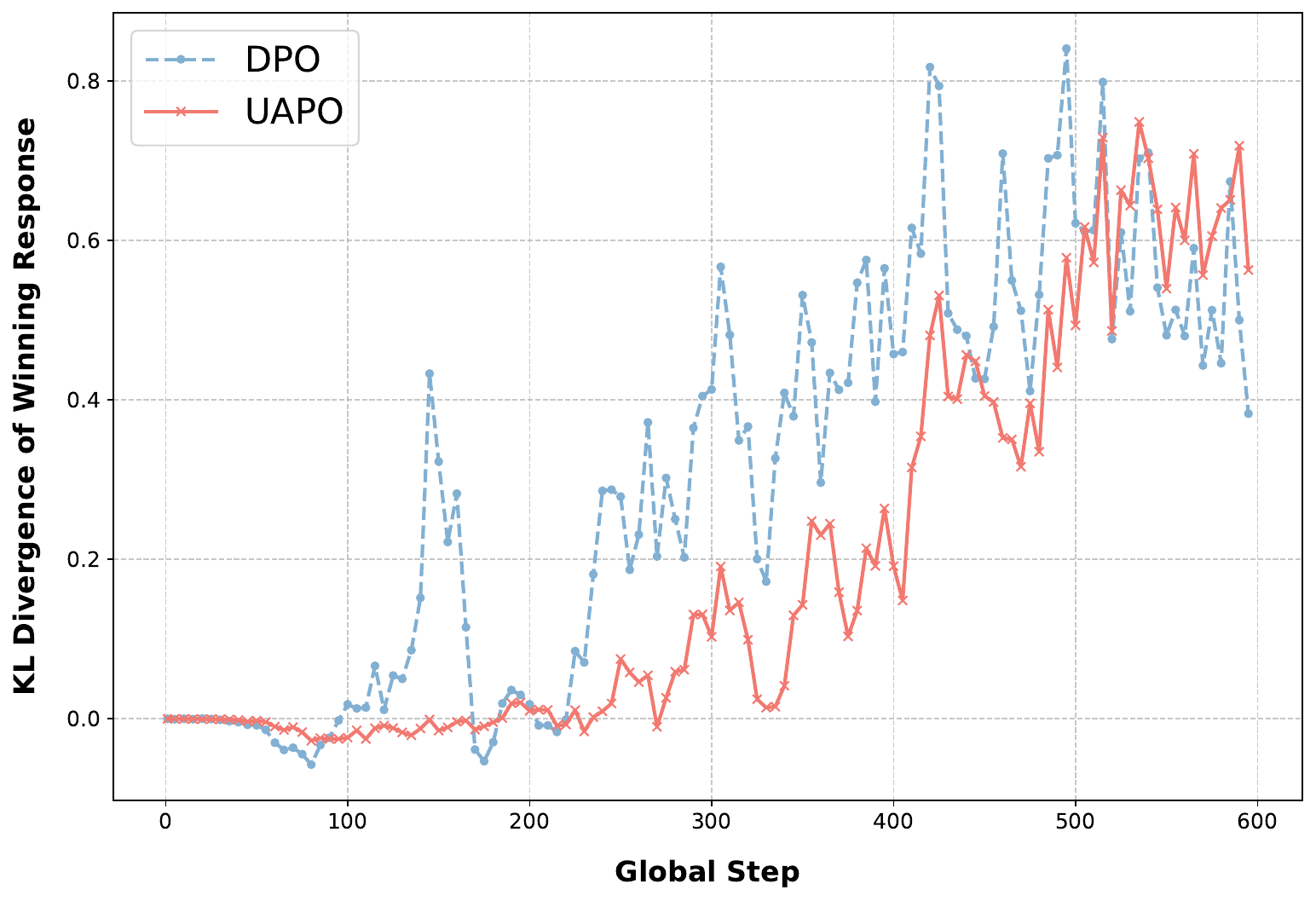}
    \label{fig:dpo-kl}
  } 
  \subfigure[SimPO vs. \SimPO]{ 
    \includegraphics[width=0.45\linewidth]{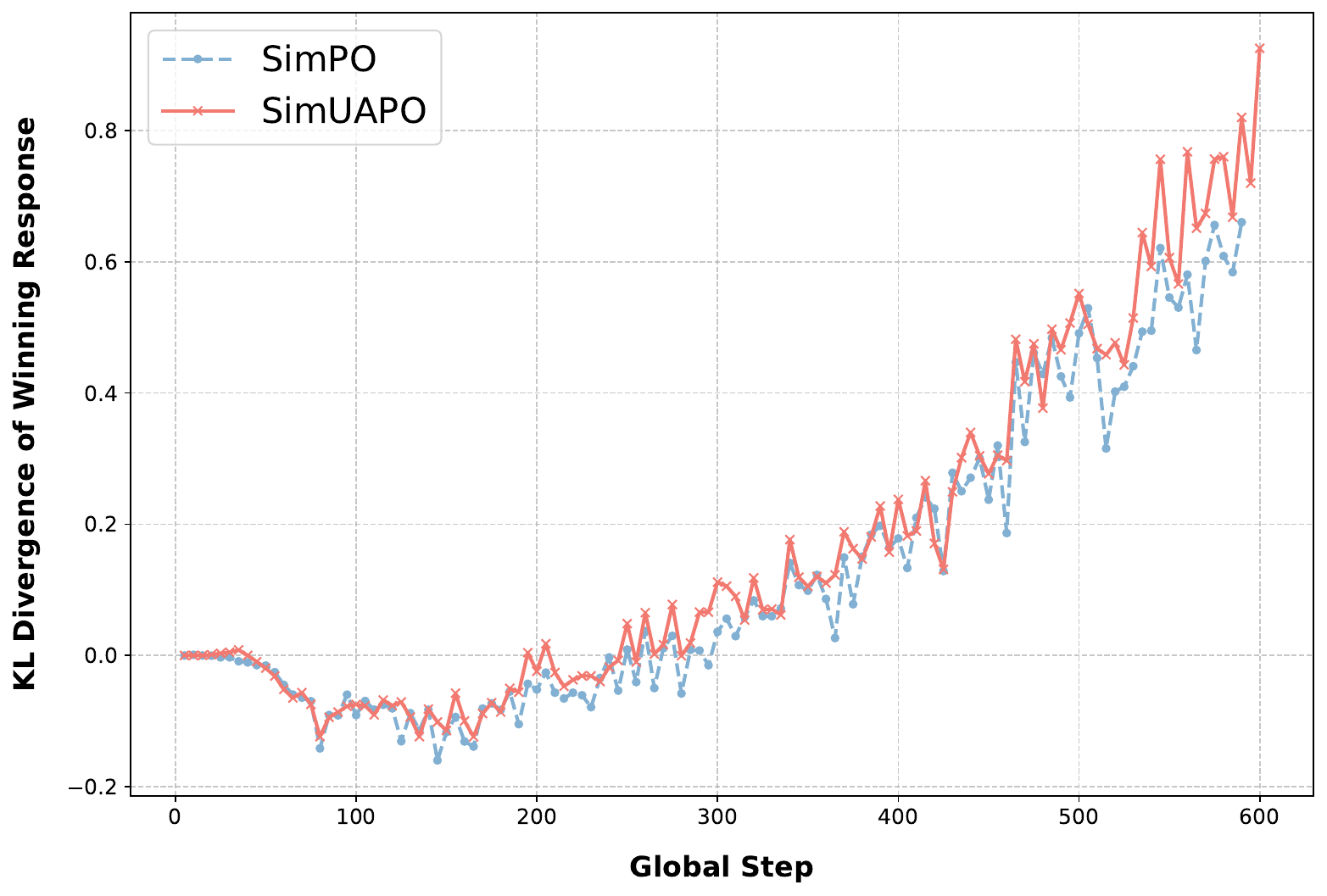}
    \label{fig:simpo-kl}
  } 
  \caption{Analysis of KL divergence on Mistral-Instruct. (a) KL divergence of winning response on DPO and \DPO over training steps, and (b) KL divergence of winning response on SimPO and \SimPO over training steps.} 
  \label{fig:KL_divergence}
\end{figure}

During training, smooth KL divergence is essential for preserving the policy model’s original behavior. As shown in \cref{fig:dpo-kl} and \cref{fig:simpo-kl}, \DPO and \SimPO exhibit lower and more stable KL divergence compared to DPO and SimPO, especially at the beginning of training. This demonstrates \DPO’s and \SimPO’s superior control over KL divergence. Overall, both methods outperform their counterparts in performance while maintaining smooth, low KL divergence throughout training.

\section{Related Work}

Aligning LLMs with human preferences is a critical area of research. As mentioned before, several lines of research have been proposed to address this challenge, broadly categorized into approaches with explicit reward models and implicit reward models.

\paragraph{Explicit reward model approaches.}
Reinforcement Learning from Human Feedback (RLHF)~\cite{ouyang2022training} is a method used to fine-tune large language models (LLMs) by incorporating human-generated feedback. In RLHF, the process typically begins with training a reward model based on human preferences, where human evaluators provide feedback on model outputs. This reward model is then used to guide the reinforcement learning process, where the LLM’s parameters are adjusted to align its outputs with the human preferences. RLHF has been shown to significantly improve the alignment of LLMs, making their responses more aligned with human values and expectations. By iterating through this feedback loop, models can be fine-tuned to better match human decision-making and moral judgments. Building on this, Reinforcement Learning with AI Feedback (RLAIF)~\cite{lee2023rlaif} extends the RLHF framework by replacing human feedback with AI-generated feedback. This innovative approach utilizes AI systems to evaluate and provide feedback on the model’s outputs, enabling alignment without the need for direct human input. RLAIF addresses some of the scalability challenges faced by RLHF, particularly in situations where large-scale human evaluation is impractical or costly. AI-generated feedback can be automated, allowing for faster iterations and potentially large-scale improvements in model behavior.

\paragraph{Implicit reward model approaches.}
Direct Preference Optimization (DPO)~\cite{rafailov2023dpo} simplifies the alignment process by directly optimizing the model based on human preferences, eliminating the need for a separate reward model. Identity Preference Optimization (IPO)~\cite{azar2024general} tried to resolve the issue of overfitting adding a regularization item. Contrastive Preference Optimization (CPO)~\cite{xu2024contrastive} focuses on contrastive learning to align models with human preferences. By contrasting preferred outputs against less preferred ones, CPO effectively guides the model toward desired behaviors. Kahneman-Tversky Optimization (KTO)~\cite{ethayarajh2024kto} relies simply on binary feedback, which are more readily obtainable and more scalable for large datasets, as it reduces the need for pairwise data. 
Odds Ratio Preference Optimization (ORPO) ~\cite{hong2024orpo} removes the need for a reference model and integrates supervised fine-tuning (SFT) with alignment into a single step. This approach simplifies the training process while maintaining alignment performance. R-DPO ~\cite{DBLP:conf/acl/rdpo} extends DPO by incorporating length normalization to disentangle the effects of response length and quality. This method helps improve the robustness of DPO in scenarios where response length may influence preference judgments.
Simple Preference Optimization (SimPO) ~\cite{meng2024simpo} is a variant of DPO that uses a reference-free reward with response length normalization. SimPO has demonstrated superior performance across diverse tasks by introducing a target reward margin to ensure a distinction between winning and losing responses. Robust DPO (rDPO) ~\cite{rdpo} proposes an unbiased loss function to mitigate the impact of known label noise in pairwise preferences, while Distributionally Robustifying DPO (Dr.DPO) ~\cite{drdpo} adopts distributionally robust optimization to reweight samples and address both pointwise and pairwise noise effectively.

\section{Conclusion}

In this paper, we propose a general framework, \method, for offline preference optimization, which introduces a novel component, the uncertainty-aware utility anchor, to enhance learning stability and effectiveness. We validate \method on both DPO and SimPO, demonstrating that it achieves strong and consistent performance across three benchmarks. Unlike traditional approaches that rely solely on pairwise preference data, \method can incorporate multiple pair data, leading to significant improvement. These findings highlight the versatility and robustness of \method, making it a promising approach for real-world preference optimization tasks, especially in settings with limited data or computational resources.


\section*{Limitations}

While \method serves as a general framework for offline preference optimization, we have only verified its performance on DPO and SimPO. Additionally, since \method introduces a utility anchor to preference optimization, it is crucial to investigate whether this anchor can further enhance the model through self-improvement. Future work could explore its application to other methods within this framework to further validate its generality.



\section*{Acknowledgments}

This work was done during an internship at State Key Laboratory of General Artificial Intelligence, BIGAI. And this work was supported by the National Natural Science Foundation of China (62376031, 62337001), the Key Technologies R \& D Program of Anhui Province (No. 202423k09020039), and the Fundamental Research Funds for the Central Universities.

\bibliography{main}

\appendix

\onecolumn

\section{Derivation of \cref{con:multi_loss}} \label{appendix:multi}

In \cref{eq:single_loss}, the sigmoid function $\sigma(\cdot)$ can be expanded, and the loss can be rewritten as:
\[
\mathcal{L}_{\text{UAPO}}(\pi_\theta) = -\mathbb{E}_{(x, y_w, y_l) \sim \mathcal{D}} \left[ 
\log \frac{\exp(r(x, y_w))}{\exp(r(x, y_\bot)) + \exp(r(x, y_w))} 
+ \log \frac{\exp(r(x, y_\bot))}{\exp(r(x, y_\bot)) + \exp(r(x, y_l))} 
\right].
\]

This expression can be decomposed into two terms:
\[
\mathcal{L}_w = - \log \left( \frac{\exp(r(x, y_w))}{\exp(r(x, y_\bot)) + \exp(r(x, y_w))} \right), 
\quad 
\mathcal{L}_l = - \log \left( \frac{\exp(r(x, y_\bot))}{\exp(r(x, y_\bot)) + \exp(r(x, y_l))} \right).
\]

When a given prompt is associated with multiple winning responses $(y_{w1}, y_{w2}, \ldots, y_{wn})$ and multiple losing responses $(y_{l1}, y_{l2}, \ldots, y_{lm})$, the reward of each winning response should exceed the utility anchor, which in turn should exceed the reward of each losing response. Under this setting, the two terms can be generalized as:
\[
\mathcal{L}_w = -\sum_{y_w} \log \left( \frac{\exp(r(x, y_w))}{\exp(r(x, y_\bot)) + \sum_{y_w'} \exp(r(x, y_w'))} \right),
\]

\[
\mathcal{L}_l = - \log \left( \frac{\exp(r(x, y_\bot))}{\exp(r(x, y_\bot)) + \sum_{y_l} \exp(r(x, y_l))} \right).
\]

Therefore, the \method objective with unpaired data can be expressed as \cref{con:multi_loss}.

\section{Derivation of \cref{eq:confidence_bound}}
\label{appendix:elbo}

A KL constrained policy optimization has been substantially defined by prior works~\cite{rafailov2023dpo}. Following prior works~\cite{jin2021pessimism}, we can define a pessimistic function with reward uncertainty as a lower confidence bound. Typically, it has a form:
\[
\max_\theta \mathbb{E}_{\mathcal{D}} \left[ r(x) - \textcolor{red}{\hat{u}(x)}  \right] - \beta \mathbb{D}_{\text{KL}} [\pi_\theta(x) || \pi_{\rm ref}(x)],
\]    
where $\hat{u}(x)$ is a uncertainty penalisation for input $x$. 

Following prior work~\cite{peters2007reinforcement}, an optimal solution to the KL-constrained reward maximization:
\[
    \pi(y|x) = \frac{1}{Z(x)} \pi_{\rm ref} (y|x) \exp \Big(\frac{1}{\beta} \left(r(x, y) - \hat{u}(x)\right)\Big).
\]

Taking the above optimal function into the Bradley-Terry from as in DPO~\cite{rafailov2023dpo}, we have:
\[
\mathcal{L}_{\rm DPO}(\pi_\theta) = -\mathbb{E}_{\mathcal{D}} \left[
\log \sigma \left( \beta \log\frac{\pi_\theta (y_w|x)}{\pi_{\rm ref} (y_w|x)} - \beta \log\frac{\pi_\theta (y_l|x)}{\pi_{\rm ref} (y_l|x)} \textcolor{red}{- \hat{u}(x) + \hat{u}(x)}\right)
\right].
\]

Taking the implicit preference objective as in \citet{rafailov2023dpo} of pair $y_w, y_l$, we have the following inequalities:
\[
\hat{r}(x, y_w) = \beta \log\frac{\pi_\theta (y_w|x)}{\pi_{\rm ref} (y_w|x)} \geq \hat{u} (x) \geq \hat{r}(x, y_l) = \beta \log\frac{\pi_\theta (y_l|x)}{\pi_{\rm ref} (y_l|x)} .
\]

Since the sigmoid function is monotonically increasing, we have
\[
\log \sigma \left(\hat{r}(x, y_w) - \hat{u}(x)\right) + \log \sigma\left(\hat{u}(x) - \hat{r}(x, y_l)\right) \leq \log \sigma\left(\hat{r}(x, y_w) - \hat{r}(x, y_l)\right),
\]
\[
\mathcal{L}_{\rm DPO}(\pi_\theta) \leq -\mathbb{E}_{\mathcal{D}} \left[
\log \sigma \left( \beta \log\frac{\pi_\theta (y_w|x)}{\pi_{\rm ref} (y_w|x)} \textcolor{red}{- \hat{u}(x) }\right) + \log \sigma \left(  \textcolor{red}{\hat{u}(x)}- \beta \log\frac{\pi_\theta (y_l|x)}{\pi_{\rm ref} (y_l|x)} \right)
\right].
\]

Therefore, \method in \cref{eq:single_loss} can be considered as a lower bound optimization of the above uncertainty-aware optimization.

\section{Practical Implementation of Utility Anchors} \label{appendix:utility_anchor}

For \DPO, we set 
\[
    \log \pi (x, y_\bot) =  \log \prod_{i=1}^n \sigma(Wh_i + b).
\]
For \SimPO, we set 
\[
    \log \pi (x, y_\bot) = \frac{1}{n} \log \prod_{i=1}^n \sigma(Wh_i + b),
\]
where $W \in \mathbb{R}^{1 \times d}$, $b \in \mathbb{R}$, $\sigma$ is the sigmoid function, $d$ is the hidden size of last layer, $n$ denotes the number of tokens in the prompt and $h_i$ denotes the output of the last hidden state corresponding to the $i$-th token of the prompt $x$. 

For the representation of $y_\bot$, we assume it depends solely on the prompt $x$. Therefore, we use the hidden states of $x$ as input to a linear function, whose output is the utility anchor. The weights of this linear function are learnable and updated during the preference optimization process, rather than being fixed in advance. This method allows the utility anchor to adjust according to the nuances of different prompts, thereby improving the alignment of the model’s outputs with human preferences.

In this paper, we use DPO and SimPO reward function to verify the feasibility of \method, \ie replace $r(x, y)$ in \cref{eq:single_loss} and \cref{con:multi_loss} with $r_{\text{DPO}}(x, y)$ or $r_{\text{SimPO}}(x, y)$, denoted as \DPO and \SimPO.

\section{Multiple Data Construction} \label{appendix:multi_data}

Following ~\cite{meng2024simpo}, we construct the multiple dataset using data from princeton-nlp/mistral-instruct-ultrafeedback, princeton-nlp/llama3-ultrafeedback-armorm, and princeton-nlp/gemma2-ultrafeedback-armorm. In these datasets, for each prompt \( x \), five responses are generated by the SFT model with a sampling temperature of 0.8. Next, the reward model annotates these responses to identify the highest-scoring one, denoted as \( y_w \), and the remaining four responses are considered as \( y_l \). To construct the multiple dataset, we select the \( y_w \) response and randomly choose three out of the four \( y_l \) responses. An example of this process can be seen in \cref{tab:multiple_dataset}.

\begin{longtable}{p{3cm}p{12cm}}
\hline
\textbf{Prompt} & how can i develop a habit of drawing daily \\
\hline
\endfirsthead
\hline
\endhead
\hline
\endfoot
\hline
\endlastfoot

\textbf{\makecell[l]{The Winning \\ Response}} & Developing a daily habit of drawing takes time and dedication, but it's a rewarding journey. Here's a breakdown of how to make it happen:

\textbf{1. Start Small and Be Realistic:} \uline{Tiny Habits:} Don't aim for masterpieces right away. Begin with 5-10 minutes a day. You'd be surprised how much you can accomplish even in short bursts. \uline{Easy Wins:} Choose simple subjects to start with: doodles, shapes, still lifes, or even tracing. Build confidence with achievable goals.

\textbf{2. Schedule It In:} \uline{Treat it like an appointment:} Choose a specific time each day dedicated to drawing. Put it in your calendar and stick to it as much as possible. \uline{Morning Ritual:} Drawing first thing in the morning can set a creative tone for the day. \uline{Evening Wind-Down:} If you're more relaxed in the evenings, drawing can be a calming way to de-stress. 

\textbf{3. Create a Conducive Environment:} \uline{Designated Space:} Set up a comfortable and inspiring drawing area. \uline{Gather Supplies:} Have your pencils, paper, erasers, etc., readily available.

\textbf{4. Remove Distractions:} \uline{Phone on Silent:} Put your phone away during your drawing time. \uline{Quiet Space:} Find a spot where you won't be interrupted.

\textbf{5. Embrace Imperfection:} \uline{Don't Edit:} Let go of the need for perfection. Focus on the process, not the outcome.\uline{Experiment:} Try different styles, tools, and subjects. Don't be afraid to make mistakes—they're part of the learning process.

\textbf{6. Find Inspiration:} \uline{Follow Artists:} Look at the work of artists you admire.\uline{Take Photos:} Capture interesting scenes or objects to draw later.\uline{Visit Museums:} Observe art in person.

\textbf{7. Join a Community:} \uline{Online Groups:} Connect with other artists for support, feedback, and inspiration.\uline{Local Classes:} Consider taking a drawing class to learn new techniques and meet fellow artists.

\textbf{8. Track Your Progress:} \uline{Sketchbook:} Keep a sketchbook to record your daily drawings. Seeing your improvement over time can be incredibly motivating.\uline{Photo Journal:} Take photos of your finished drawings to document your journey.

\textbf{9. Reward Yourself:} \uline{Celebrate Milestones:} Treat yourself when you reach a drawing goal.\uline{Enjoy the Process:} The most important reward is the joy of creating!

Remember, consistency is key. Even if you only draw for a few minutes each day, you'll see progress over time. Be patient, be persistent, and most importantly, have fun! \\

\textbf{\makecell[l]{The First \\ Losing Response}} & Developing a daily drawing habit takes dedication and a sprinkle of strategy. Here's a roadmap to help you on your artistic journey:

\textbf{1. Set Realistic Goals:} \uline{Start Small:} Don't pressure yourself to create masterpieces every day. Begin with 15-20 minutes of drawing. \uline{Focus on Consistency:} Aim for daily practice, even if it's just quick sketches. It's about building the habit, not producing perfect artwork.

\textbf{2. Make it Enjoyable:} \uline{Choose Subjects You Love:} Draw things that genuinely interest you – people, animals, landscapes, objects, patterns – anything that sparks your creativity. \uline{Experiment with Mediums:} Don't limit yourself to pencils. Explore pens, markers, watercolors, digital tools, or even charcoal.\uline{Listen to Music or Podcasts:} Create a soundtrack or listen to something engaging while you draw to keep your mind stimulated.

\textbf{3. Create a Routine:} \uline{Schedule It In:} Treat your drawing time like an important appointment. Block it out on your calendar and stick to it as much as possible.\uline{Find Your Ideal Time:} Experiment with different times of day to see when you're most creative.\uline{Set Up a Dedicated Space:} Even a small corner of your room can become your art haven. Keep your drawing supplies organized and easily accessible.

\textbf{4. Embrace the Process:} \uline{Don't Be Afraid to Make Mistakes:} Mistakes are part of learning. Embrace them as opportunities to grow. \uline{Focus on Progress, Not Perfection:} Celebrate small victories and track your improvement over time.\uline{Learn from Others:} Study the work of artists you admire, take online courses, or join a drawing group for inspiration and feedback.

\textbf{5. Stay Motivated:} \uline{Join a Challenge:} There are many online drawing challenges that can provide structure and accountability.\uline{Share Your Work:} Show your drawings to friends, family, or online communities. Feedback and encouragement can be incredibly motivating. \uline{Reward Yourself:} Celebrate milestones with small rewards to keep yourself on track. 

\textbf{Remember:} Developing a daily habit takes time and effort. Be patient with yourself, stay consistent, and most importantly, have fun! \\

\textbf{\makecell[l]{The Second \\ Losing Response}} & It's awesome that you want to make drawing a daily habit! It's a rewarding journey. Here's a breakdown of how to make it stick:

\textbf{1. Start Small \& Be Realistic:} \uline{Don't aim for masterpieces:} Begin with 5-10 minutes a day. Even doodling is progress! \uline{Set achievable goals:} Instead of "draw for an hour," try "sketch something for 15 minutes." \uline{Be kind to yourself:} Some days will be better than others. Don't get discouraged if you miss a day or two.

\textbf{2. Make it Enjoyable:} \uline{Choose subjects you love:} Draw things that excite you, whether it's animals, landscapes, people, or abstract shapes. \uline{Experiment with different mediums:} Explore pencils, pens, charcoal, watercolors, digital tools – find what feels right. \uline{Listen to music or podcasts:} Create a relaxing or inspiring atmosphere while you draw.

\textbf{3. Build a Routine:} \uline{Pick a specific time:} Morning, lunch break, evening – find a time that works best for you and stick to it. \uline{Set reminders:} Use your phone, calendar, or sticky notes to keep yourself on track. \uline{Create a dedicated space:} Having a designated drawing area can signal to your brain that it's time to create.

\textbf{4. Stay Inspired:} \uline{Carry a sketchbook:} Jot down ideas, quick sketches, or observations throughout the day. \uline{Follow artists you admire:} Look at their work, learn from their techniques, and get inspired. \uline{Take online classes or workshops:} Learn new skills and challenge yourself.

\textbf{5. Embrace the Process:} \uline{Focus on the journey, not just the destination:} Enjoy the act of creating, even if the final result isn't perfect. \uline{Don't be afraid to make mistakes:} They're opportunities to learn and grow. \uline{Celebrate your progress:} Track your drawings, look back on how far you've come, and be proud of your dedication.

Remember, the key is to be consistent and patient with yourself. Enjoy the process of learning and growing as an artist! \\

\textbf{\makecell[l]{The Third \\ Losing Response}} & \textbf{How to Develop a Daily Drawing Habit}

It's fantastic that you want to make drawing a daily practice! Here's a roadmap to help you build that habit:

\textbf{1. Start Small and Be Realistic:} \uline{Don't aim for masterpieces:} Begin with 5-10 minutes a day. Focus on consistency over quantity or quality initially. \uline{Set achievable goals:} Instead of "draw for an hour," try "draw for 15 minutes before breakfast."

\textbf{2. Make it Enjoyable:} \uline{Choose subjects you love:} Draw things that spark your interest, whether it's nature, people, animals, or abstract shapes. \uline{Experiment with different mediums:} Don't limit yourself to pencils. Try pens, charcoal, watercolors, digital tools – explore and find what you enjoy most. \uline{Listen to music or podcasts:} Find something that helps you relax and focus while you draw.

\textbf{3. Create a Routine:} \uline{Schedule it in:} Treat drawing like any other important appointment. Block out a specific time in your day and stick to it. \uline{Set up your space:} Have your drawing materials easily accessible in a dedicated area. \uline{Use a timer:} Knowing you only have a set amount of time can help you get started and stay focused.

\textbf{4. Embrace Imperfection:} \uline{Don't be afraid to make mistakes:} Mistakes are part of the learning process. Embrace them and learn from them. \uline{Don't compare yourself to others:} Everyone progresses at their own pace. Focus on your own journey and celebrate your progress.

\textbf{5. Find Inspiration:} \uline{Follow artists you admire:} Observe their techniques and styles. \uline{Visit museums or galleries:} Expose yourself to different art forms and perspectives. \uline{Draw from life:} Observe the world around you and capture what you see.

\textbf{6. Track Your Progress and Celebrate Milestones:} \uline{Keep a sketchbook:} Document your daily drawings and track your progress over time. \uline{Share your work:} Show your drawings to friends, family, or online communities for feedback and encouragement. \uline{Reward yourself:} Acknowledge your achievements and celebrate your dedication.

\textbf{Remember:} Building a daily habit takes time and effort. Be patient with yourself, stay consistent, and enjoy the process!

Let me know if you have any other questions. \\

\label{tab:multiple_dataset}

\end{longtable}

\section{Experiments}

\subsection{The robustness of \method over different constant reward anchors}

We also evaluate the performance of \SimPO and SimPO under different constant reward anchors (\( \gamma \)) settings on the Gemma-2-Instruct model in \cref{tab:margin} \footnote{- denotes that the model has collapsed after training.}. Across various constant reward anchors, \SimPO consistently maintains high performance on AlpacaEval 2, demonstrating its robustness to changes in hyperparameters. In contrast, SimPO only achieves satisfactory performance when \( \gamma = 5 \), which fails to generate coherent sequences under other constant reward anchor settings. 

\begin{table}[H]
    \centering
    \scalebox{0.87}{
    \begin{tabular}{ccccc}
        \toprule
        \multirow{2}{*}{\textbf{constant reward anchor $\gamma$}} & \multicolumn{2}{c}{\textbf{\SimPO}} & \multicolumn{2}{c}{\textbf{SimPO}} \\
        \cmidrule(lr){2-3}  \cmidrule(lr){4-5} 
         & LC (\%) & WR (\%) & LC (\%) & WR (\%) \\
        \midrule
        5   & 67.7  & 62.3 & \textbf{72.4} & \textbf{65.9} \\
        10 & 70.7 & 65.1  & - & - \\
        16   & \textbf{73.5}  & \textbf{67.0} & - & - \\
        20 & 72.2  & 65.0  & - & - \\
        \bottomrule
    \end{tabular}}
    \caption{Performance of using different constant reward anchors on Gemma-2-Instruct.} 
    \label{tab:margin}
    \vspace{-8pt}
\end{table}

This performance disparity highlights the robustness of \SimPO compared to SimPO. The key factor that contributes more is the utility anchor in \SimPO, which effectively regulates the KL divergence between the policy model and the reference model. By dynamically adjusting the utility anchor, it is more robust to \SimPO varying with hyperparameters changes (such as \( \gamma \)). This adaptability prevents excessive divergence from the reference model, which is more consistent and reliable performance across different training configurations.

\subsection{The statistical significance testing of \method}

To evaluate statistical significance, we conduct three evaluation runs on Mistral-Instruct using AlpacaEval 2 for SimPO, \SimPO, and \SimPO-multi, as shown in \cref{tab:statistical}. The results show that \method is stable, as evidenced by its low standard deviation and notably low coefficient of variation, and that \SimPO-multi consistently outperforms SimPO by a significant margin in terms of win rate.

\begin{table}[H]
\centering
\scalebox{0.87}{
\begin{tabular}{lcccc}
\hline
\textbf{Method} & \textbf{AlpacaEval 2 LC (\%)} & \textbf{CV} & \textbf{AlpacaEval 2 WR (\%)} & \textbf{CV} \\
\hline
SimPO & 28.868 $\pm$ 0.147 & 0.509\% & 31.837 $\pm$ 0.131 & 0.411\% \\
SimUAPO &  28.870$\pm$ 0.223  & 0.774\% & 32.257 $\pm$ 0.021 & 0.065\% \\
SimUAPO-multi & 22.367 $\pm$ 0.610 & 2.727\% & 40.463 $\pm$ 0.254 & 0.628\% \\
\hline
\end{tabular}}
\caption{The statistical significance testing results on AlpacaEval 2 with corresponding coefficient of variation (CV) on Mistral-Instruct.}
\label{tab:statistical}
\end{table}

\subsection{Downstream tasks evaluation}


We evaluate Mistral-Instruct and Llama-3-Instruct models trained with different methods on a range of tasks using the lm-evaluation-harness \footnote{https://github.com/EleutherAI/lm-evaluation-harness} ~\cite{eval-harness}. These tasks include MMLU ~\cite{mmlu}, TruthfulQA ~\cite{truthfulqa}, HellaSwag ~\cite{hellaswag}, and GSM8K ~\cite{gsm8k} and the results are shown in \cref{tab:downstream}.

\paragraph{Knowledge is effectively retained.} Both \DPO and \SimPO preserve general knowledge as measured by MMLU, with minimal degradation compared to the SFT baseline. This suggests that the introduction of utility-aware preference objectives does not compromise the model’s foundational knowledge.

\paragraph{Truthfulness improves significantly.} On the TruthfulQA benchmark, \SimPO achieves the highest scores among all methods on both Mistral-Instruct and Llama-3-Instruct models, indicating that its preference optimization process emphasizes factual consistency and helps generate more truthful responses.

\paragraph{Commonsense reasoning remains competitive.} For HellaSwag, both \DPO and \SimPO maintain performance on par with or slightly better than SFT. This suggests that utility-aware optimization does not hinder the model’s ability to perform in-context reasoning and may benefit from exposure to diverse prompts in the preference dataset.

\paragraph{Math performance is more stable.} Unlike many other methods that show sharp performance drops on GSM8K, \DPO demonstrates relatively stable results, suggesting that utility-aligned optimization may help balance the trade-off between human preference alignment and mathematical reasoning.

\paragraph{Overall performance is robust.} \method consistently delivers competitive or improved average performance across benchmarks. And it balances strong gains in truthfulness and reasoning with only modest compromises in other areas, making it a strong candidate for robust preference alignment.

\begin{table}[t!]
\centering
\resizebox{\linewidth}{!}{
\begin{tabular}{lcccccccccc}
\toprule
\multirow{2}{*}{\textbf{Method}} & \multicolumn{5}{c}{\textbf{Mistral-7B-Instruct}} & \multicolumn{5}{c}{\textbf{Llama-3-8B-Instruct}} \\
\cmidrule(lr){2-6} \cmidrule(lr){7-11}
& MMLU(0) & TruthfulQA(0) & HellaSwag(0) & GSM8K(5) & Average & MMLU(0) & TruthfulQA(0) & HellaSwag(0) & GSM8K(5) & Average \\
\midrule
SFT     & 59.06 & 66.87 & 83.67 & 42.23 & 62.96 & 63.79 & 51.56 & 75.88 & 75.28 & 66.63   \\
DPO     & 59.27 & 66.74 & 84.53 & 42.76 & 63.33 & 64.61 & 55.50 & 76.23 & 76.42 & 68.19   \\
SimPO   & 56.62 & 71.47 & 85.32 & 22.06 & 58.87 & 64.40 & 64.09 & 75.29 & 56.56 & 65.09   \\
UAPO    & 57.78 & 65.45 & 83.39 & 30.17 & 59.20 & 64.31 & 55.25 & 76.34 & 76.04 & 68.00   \\
SimUAPO & 57.88 & 70.04 & 84.22 & 30.63 & 60.69 & 64.32 & 62.87 & 71.89 & 71.57 & 67.66   \\
\bottomrule
\end{tabular}}
\caption{Downstream task evaluation results of tasks on the huggingface open leaderboard.}
\label{tab:downstream}
\end{table}

\subsection{The example which shows the randomness of MT-Bench} \label{appendix:mt-bench}

During our evaluation process, we frequently observe that the same model can receive varying scores on a single MT-Bench sample (e.g., see \cref{tab:mt-bench}). This variability highlights concerns regarding the consistency and reliability of MT-Bench. Consequently, we consider AlpacaEval 2 and Arena-Hard to offer more stable and trustworthy evaluations for comparing different preference optimization methods. Nevertheless, since MT-Bench remains widely used in prior work, we include it in our evaluation to ensure consistency and comparability across studies.

\begin{table}[t!]
    \centering
    \scalebox{0.6}{
    \begin{tabular}{lc}
        \toprule
       \textbf{Judgement} & \textbf{Score} \\
        \midrule
        The assistant's response to the second question contains an error. It suggests that a solution ... The corrected rating for the assistant's response would be [[5]]. & 5 \\
        The assistant's response is inaccurate. Even though the assistant's code meets the user's requirement of ... Therefore, I would rate this response as: [[4]].  & 4 \\
        \bottomrule
    \end{tabular}}
    \caption{Inconsistencies in MT-Bench scores for the same model on a single sample.}
    \label{tab:mt-bench}
    \vspace{-8pt}
\end{table}

\subsection{Multiple datasets performance on Mistral-Instruct}

We also train DPO, SimPO, \DPO, and \SimPO on Mistral-Instruct using multiple datasets and the results are in \cref{tab:mistral-multi}. \SimPO-multi achieves a 40.5 WR on AlpacaEval 2 and a 23.1 WR on Arena-Hard which are much higher than other preference optimization methods. Interestingly, on Mistral-Instruct, \SimPO-multi further enlarges the gap between LC and WR on AlpacaEval 2 compared to both SimPO and \SimPO. This discrepancy is not observed on Llama-3-Instruct, which may be attributed to the inherent characteristics of the Mistral-Instruct model and the specific design of the SimPO reward function.

\begin{table}[H]
    \centering
    \scalebox{0.8}{
    \begin{tabular}{lcccc}
        \toprule
        \multirow{2}{*}{\textbf{Method}} & \multicolumn{2}{c}{\textbf{AlpacaEval 2}} & \textbf{Arena-Hard} & \textbf{MT-Bench} \\
        \cmidrule(lr){2-3} \cmidrule(lr){4-4} \cmidrule(lr){5-5} 
        & LC (\%) & WR (\%) & WR (\%) & GPT-4 \\
        \midrule
        \multicolumn{5}{c}{\cellcolor{lightgray!50} \textbf{multiple datasets}}\\
        DPO   & 20.5  & 18.1  & 13.4 & \uline{7.6} \\
        SimPO & \textbf{29.0}  & 31.9  & \uline{19.9} & 7.2 \\ 
        \DPO & 23.1 & 17.7 & 11.1 & \textbf{7.7} \\
        \SimPO & \uline{28.6} & \uline{32.2} & 17.7 & \uline{7.6} \\
        \midrule
        DPO-multi & 17.0 & 16.7 & 8.4 & 6.8 \\
        SimPO-multi & 14.5 & 22.2 & 9.2 & 6.3 \\
        \DPO-multi & 24.1 & 18.6 & 13.8 & \textbf{7.7} \\
        \SimPO-multi & 21.9  & \textbf{40.5} & \textbf{23.1} & \uline{7.6} \\
        \bottomrule
    \end{tabular}}
    \caption{Performance comparison of different methods on Mistral-Instruct trained with multiple datasets.}
    \label{tab:mistral-multi}
\end{table}

\subsection{The necessity of constant reward anchor in \method}

To further investigate the significance of the constant reward anchor in \method, we compare the performance of \DPO and \SimPO with and without the constant reward anchor $\gamma$ in \cref{tab:margin_ablation}. The results show that removing the constant reward anchor leads to a significant performance decline for both \DPO and \SimPO across most benchmarks, underscoring its critical role in the \method framework. However, \DPO exhibits inconsistent trends without the reward anchor: while performance drops on AlpacaEval 2, it improves on Arena-Hard. This inconsistency across benchmarks when evaluating the same model highlights the need for more reliable and robust evaluation benchmarks for preference optimization.

\begin{table}[H]
    \centering
    \scalebox{0.8}{
    \begin{tabular}{lccccccccc}
        \toprule
        \multirow{3}{*}{\textbf{Method}} & \multicolumn{4}{c}{\textbf{Mistral-7B-Instruct}} & \multicolumn{4}{c}{\textbf{Gemma-2-9B-Instruct}} \\
        \cmidrule(lr){2-5} \cmidrule(lr){6-9} 
        & \multicolumn{2}{c}{\textbf{AlpacaEval 2}} & \textbf{Arena-Hard} & \textbf{MT-Bench} & \multicolumn{2}{c}{\textbf{AlpacaEval 2}} & \textbf{Arena-Hard} & \textbf{MT-Bench}  \\
        \cmidrule(lr){2-3} \cmidrule(lr){4-4} \cmidrule(lr){5-5} \cmidrule(lr){6-7} \cmidrule(lr){8-8} \cmidrule(lr){9-9} 
        & LC (\%) & WR (\%) & WR (\%) & GPT-4 &  LC (\%) & WR (\%) & WR (\%) & GPT-4 \\
        \midrule
         \multicolumn{9}{c}{\cellcolor{lightgray!50} \textbf{w constant reward anchor}}\\
        \DPO & 23.1  & 17.7 & 11.1 & \textbf{7.7} & \uline{70.2}  & \textbf{67.7} & \uline{58.8}   & \textbf{8.9}\\
        \SimPO & \textbf{28.6}  & \textbf{32.2} & \uline{17.7} & \uline{7.6}  &\textbf{73.5}  & \uline{67.0} & \textbf{59.4} & \textbf{8.9} \\
        \midrule
         \multicolumn{9}{c}{\cellcolor{lightgray!50} \textbf{w/o constant reward anchor}}\\
        \DPO & 17.6 & 14.8 & 14.9 & \uline{7.6} & 55.4 & 48.7 & 41.4 & 8.4 \\
        \SimPO & \uline{24.8} & \uline{26.9} & \textbf{22.7} & \uline{7.6} & 60.8 & 56.6 & 54.9 & \uline{8.5} \\
        \bottomrule
    \end{tabular}}
    \caption{An ablation study of the constant reward anchor-free setting on Mistral-Instruct and Gemma-2-Instruct.}
    \label{tab:margin_ablation}
\end{table}

\subsection{Generalization capability of \method on R-DPO}

To further demonstrate the generality of \method, we conduct an additional experiment by applying it to R-DPO on LLama-3-Instruct. As shown in \cref{tab:r-dpo}, R-UAPO achieves a 2.2-point improvement over R-DPO on AlpacaEval 2 LC. These results highlight that \method exhibits strong generalization capabilities.

\begin{table}[H]
\centering
\scalebox{0.87}{
\begin{tabular}{lcccc}
\hline
\textbf{Method} & \textbf{AlpacaEval 2 LC (\%)} & \textbf{AlpacaEval 2 WR (\%)} \\
\hline
R-DPO & 48.0 & 45.8 \\
R-UAPO & 50.2 & 47.1 \\
\hline
\end{tabular}}
\caption{The performance of R-DPO and R-UAPO on Llama-3-Instruct trained with the pairwise dataset.}
\label{tab:r-dpo}
\end{table}

\section{Implementation Details} 

\subsection{Training Hyperparameters}

Since the performance of most offline preference optimization methods is highly sensitive to training hyperparameters, such as learning rate, batch size, and others, we conducted a hyperparameter search, as specified in the respective papers, to ensure a fair comparison. The results are summarized in \cref{tab:hyperparameters}. We use a batch size of 64 for training all methods. The learning rate is selected from the range \( [3e-7, 5e-7, 7e-7, 1e-6] \). All models are trained for one epoch using the Adam optimizer~\cite{DBLP:journals/corr/adam}.

\begin{table}[h]
\centering
\resizebox{\linewidth}{!}{
\begin{tabular}{@{}ll@{\hspace{1em}}l}
\toprule
\textbf{Method} & \textbf{Objective} & \textbf{Hyperparameter} \\ 
\midrule
DPO ~\cite{rafailov2023dpo}  & $-\log \sigma \left( \beta \log \frac{\pi_\theta(y_w|x)}{\pi_{\text{ref}}(y_w|x)} - \beta \log \frac{\pi_\theta(y_l|x)}{\pi_{\text{ref}}(y_l|x)} \right)$ & $\beta \in [0.01, 0.05, 0.1]$ \\
\midrule
IPO ~\cite{azar2024general}  & $\left( \log \frac{\pi_\theta(y_w|x)}{\pi_{\text{ref}}(y_w|x)} - \log \frac{\pi_\theta(y_l|x)}{\pi_{\text{ref}}(y_l|x)} - \frac{1}{2\tau} \right)^2$ & $\tau \in [0.01, 0.1, 0.5, 1.0]$ \\
\midrule
CPO ~\cite{xu2024contrastive} & $-\log \sigma \left( \beta \log \pi_\theta(y_w|x) - \beta \log \pi_\theta(y_l|x) \right) - \lambda \log \pi_\theta(y_w|x)$ & $\lambda = 1.0, \beta \in [0.01, 0.05, 0.1]$ \\
\midrule
\multirow{2}{*}{KTO ~\cite{ethayarajh2024kto}}  & $-\lambda_w \sigma \left( \beta \log \frac{\pi_\theta(y_w|x)}{\pi_{\text{ref}}(y_w|x)} - z_{0} \right) + \lambda_l \sigma \left( z_{0} - \beta \log \frac{\pi_\theta(y_l|x)}{\pi_{\text{ref}}(y_l|x)} \right)$, & $\lambda_l = \lambda_w = 1.0$ \\ 
& where $z_{0} = \mathbb{E}_{(x, y) \sim \mathcal{D}} \left[\beta\text{KL}\left(\pi_\theta(y|x) \| \pi_{\text{ref}}(y|x)\right)\right]$ & $\beta \in [0.01, 0.05, 0.1]$ \\
\midrule
\multirow{2}{*}{ORPO ~\cite{hong2024orpo}} & $-\log p_\theta(y_w|x) - \lambda \log \sigma \left( \frac{p_\theta(y_w|x)}{1 - p_\theta(y_w|x)} \right) - \log \left( \frac{p_\theta(y_l|x)}{1 - p_\theta(y_l|x)} \right)$, &  \multirow{2}{*}{$\lambda \in [0.1, 0.5, 1.0, 2.0]$}\\ 
& where $p_\theta(y|x) = \exp \left( \frac{1}{|y|} \log \pi_\theta(y|x) \right)$ & \\
\midrule
\multirow{2}{*}{R-DPO ~\cite{DBLP:conf/acl/rdpo}} & \multirow{2}{*}{$-\log \sigma \left( \beta \log \frac{\pi_\theta(y_w|x)}{\pi_{\text{ref}}(y_w|x)} - \beta \log \frac{\pi_\theta(y_l|x)}{\pi_{\text{ref}}(y_l|x)} + \left(\alpha |y_w| - \alpha |y_l|\right) \right)$} & $\alpha \in [0.05, 0.1, 0.5, 1.0]$ \\
& & $\beta \in [0.01, 0.05, 0.1]$\\
\midrule
\multirow{2}{*}{SimPO ~\cite{meng2024simpo}} & \multirow{2}{*}{$-\log \sigma \left( \frac{\beta}{|y_w|} \log \pi_\theta(y_w|x) - \frac{\beta}{|y_l|} \log \pi_\theta(y_l|x) - \gamma \right)$} & $\beta \in [2.0, 2.5, 10.0]$ \\
& & $\gamma \in [0.3, 1.0, 1.6, 3.0, 5.0]$\\
\midrule
\DPO & $-\log \sigma \Big( r_{\text{\DPO}} (x, y_w) - r_{\text{\DPO}}(x, y_{\bot}) \Big) - \log \sigma \Big( r_{\text{\DPO}}(x, y_{\bot}) - r_{\text{\DPO}} (x, y_l) \Big)$ & $\beta \in [0.01, 0.05]$, $\gamma \in [1.0, 4.5, 8.0]$\\ 
\midrule
\SimPO & $-\log \sigma \Big( r_{\text{\SimPO}} (x, y_w) - r_{\text{\SimPO}}(x, y_{\bot}) \Big) - \log \sigma \Big( r_{\text{\SimPO}}(x, y_{\bot}) - r_{\text{\SimPO}} (x, y_l) \Big)$ & $\beta \in [2.5, 10.0]$, $\gamma \in [4.5, 8.0]$\\ 
\bottomrule
\end{tabular}}
\caption{The objectives and hyperparameters of different offline preference optimization methods.}
\label{tab:hyperparameters}
\end{table}

\subsection{Computing Resources}

All the experiments can be conducted on eight Nvidia A100 80GB GPUs, 32GB memory, and a 128-core AMD CPU.

\subsection{Training Time}

Since \method introduces additional computational overhead for optimizing and inferring dummy tokens, we compare the training time of different methods. All experiments are conducted with PyTorch 2.3.0 on Ubuntu 18.04. The results show the training time required for each method in \cref{tab:training-time}.

\begin{table}[H]
\centering
\scalebox{0.8}{
\begin{tabular}{lccc}
\toprule
\textbf{Method} & \textbf{Mistral-7B-Instruct} & \textbf{Llama-3-8B-Instruct} & \textbf{Gemma-2-9B-Instruct} \\
\midrule
DPO     & 5.3h & 5.0h & 4.7h \\
SimPO   & 4.7h & 5.3h & 10.9h \\
\DPO    & 5.9h & 5.3h & 5.1h \\
\SimPO & 3.3h & 6.0h & 12.1h \\
\bottomrule
\end{tabular}}
\caption{Training time comparison across different methods.}
\label{tab:training-time}
\end{table}

\clearpage

\end{document}